\documentclass[11pt,a4paper]{article}

\usepackage[a4paper,margin=2.2cm]{geometry}
\usepackage{microtype}
\usepackage{amsmath,amssymb,mathtools}
\usepackage{booktabs}
\usepackage{graphicx}
\usepackage{placeins}
\usepackage{xcolor}
\usepackage{tikz}
\usetikzlibrary{arrows.meta,positioning,fit,backgrounds,decorations.pathmorphing}
\usepackage{pgfplots}

\pgfplotsset{compat=1.18}
\usepackage{newunicodechar}
\newcommand*\circled[1]{\tikz[baseline=(char.base)]{\node[shape=circle,draw,inner sep=0.75pt,line width=0.45pt,font=\normalfont\scriptsize] (char) {#1};}}
\newunicodechar{①}{\circled{1}}
\newunicodechar{②}{\circled{2}}
\newunicodechar{③}{\circled{3}}
\newunicodechar{④}{\circled{4}}
\newunicodechar{⑤}{\circled{5}}
\usepackage[colorlinks=true,allcolors=blue]{hyperref}
\usepackage{xurl}
\usepackage[numbers,sort&compress]{natbib}

\title{A Physical Response-and-Memory Model for Muon Optimization}
\author{Yinze Hu$^{1,2}$, Hongjun Xiang$^{1}$, Xingao Gong$^{1}$, Hongyu Yu$^{1,}$\thanks{Corresponding author: hongyuyu20@fudan.edu.cn}\\[6pt]
{\small\itshape $^{1}$Key Laboratory of Computational Physical Sciences (Ministry of Education), Institute of Computational}\\
{\small\itshape Physical Sciences, State Key Laboratory of Surface Physics, and Department of Physics,}\\
{\small\itshape Fudan University, Shanghai 200433, China}\\
{\small\itshape $^{2}$School of Physics Science and Engineering, Tongji University, Shanghai 200092, China}}
\date{}
\begin{document}
\maketitle

\begin{abstract}
\noindent
The Muon optimizer performs strongly in practice, but why its semi-orthogonalized update works and how long a history its momentum should retain lack a physical explanation. Here we model each weight matrix as a responsive medium with memory, from which both components follow directly. Semi-orthogonalization gives the update direction that dissipates loss fastest under a per-step output-perturbation bound, while momentum is an internal stress that relaxes exponentially. Because both reflect the simplest idealized medium, we obtain PhysMuon by calibrating output bounds to channel-wise gradient fluctuations and combining fast and slow relaxation modes. PhysMuon reaches the target loss earlier than tuned Muon on the GPT-2 benchmark, lowers force errors by 7--21\% across interatomic-potential tasks, and cuts unstable \(2100\,\mathrm K\) molecular-dynamics trajectories from 66.7\% to 48.8\%. On Matbench Discovery, PhysMuon raises DPA4-Pro's combined performance score from \(0.865\) to \(0.876\) and establishes the model as a state-of-the-art MPtrj-trained universal atomistic potential.
\end{abstract}

\label{pm:intro}
Training large-scale neural networks is computationally expensive; how low the loss can be driven under a fixed compute budget depends on how an optimizer converts each step's gradient into a weight update. Muon first applies an exponential moving average to past gradients to form a momentum matrix, then updates weights with its polar-decomposition orthogonal factor for semi-orthogonalization\cite{jordan2024muon}. It has repeatedly set records on a public language-model training-speed benchmark\cite{moddednanogpt}, scaled to billion-parameter models\cite{liu2025muonscalable}, and operated stably in frontier trillion-token training\cite{bai2025k2}.

Yet a clear physical picture of why these two operations work and how they should be coordinated remains lacking; in practice, how long a gradient history momentum should retain is largely set by empirical trial and error. Mathematically, Bernstein and Newhouse showed that without the exponential moving average, Muon's semi-orthogonalized direction is steepest descent under the spectral norm\cite{bernstein2024anthology,pethick2025lmo}. However, their derivation omits momentum, and they noted that the role of the exponential moving average remains open.

Subsequent modifications to Muon, such as per-neuron scaling, second-moment injection, and distributed orthogonalization\cite{normuon2025,adamuon2025,ahn2025dion,xu2026fismo,zhang2026namo,dolatabadi2026numuon,grishina2025cans,zhou2026pisa}, mostly target single-step update direction and magnitude, while the weights momentum assigns to historical gradients, the memory kernel, remain a single exponential decay. For the memory kernel, the closest works are AdEMAMix\cite{pagliardini2024ademamix}, which mixes a slow exponential moving average with Adam's fast branch, and Admeta\cite{chen2023admeta}, which uses a double exponential moving average variant in momentum optimizers. Schedule-Free also maintains a slow average over parameters\cite{morwani2025connections,defazio2024schedulefree}, but replaces learning-rate schedules rather than improving the gradient memory kernel. Although physics has widely informed neural-network analysis and design\cite{hopfield1982,bahri2020statmech,raissi2019pinn}, few works directly model training optimization\cite{deluca2022ecd,lu2026specmuon}.

To address this gap, we model each layer's weight matrix as a responsive medium with memory, where output-target mismatch exerts a gradient driving force and the weight update is the medium's response that dissipates mismatch energy (Fig.~\ref{fig:model}). Both components of standard Muon follow directly when this medium takes its simplest idealized form: semi-orthogonalization maximizes dissipation under an isotropic output-perturbation budget, while momentum is an internal stress with a single relaxation time. Generalizing these minimal assumptions to forms closer to a real medium yields PhysMuon (Fig.~\ref{fig:physmuon}), which calibrates the output budget to channel-wise gradient fluctuations and combines fast and slow relaxation modes with dynamic readout. Across language-model training on modded-nanogpt~\cite{moddednanogpt}, five machine-learned interatomic-potential tasks, Matbench Discovery, and high-temperature molecular dynamics (MD) simulations, PhysMuon consistently outperforms Muon-family baselines in convergence speed, force and energy accuracy, materials-discovery metrics, and simulation stability. These results show that optimizers can be derived and improved from the properties of a responsive medium.

\section*{Results}\label{pm:results}

\subsection*{Deriving Muon from a responsive medium}

\begin{figure}[!t]
\centering
\includegraphics[width=\textwidth]{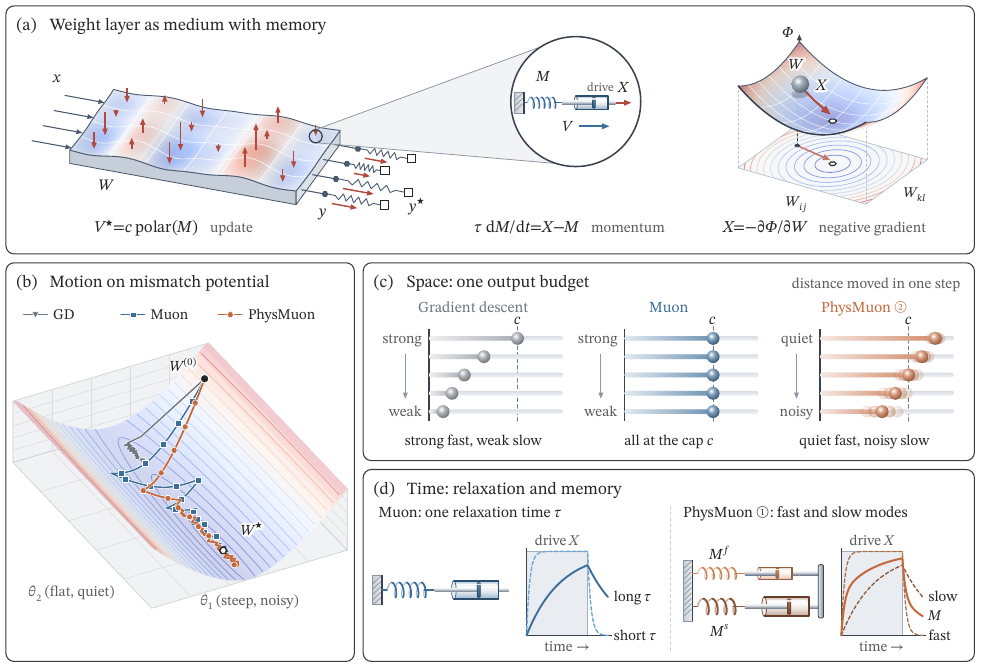}
\caption[Physical model of a responsive medium]{Physical model of a responsive medium (schematic). (a) Weight layer modelled as a responsive medium with memory: driving force \(X=-\partial\Phi/\partial W\) from mismatch potential \(\Phi\) is the negative gradient, rearrangement velocity \(V=\mathrm dW/\mathrm dt\) is the update, and internal stress \(M\) corresponds to momentum and relaxes via \(\tau\,\mathrm dM/\mathrm dt=X-M\). The memory update \(V^\star=c\,\mathrm{polar}(M)\) reduces to equation~\eqref{eq:polar} when \(M=X\). (b) Motion on the mismatch potential for gradient descent (GD), Muon, and PhysMuon. (c) Spatial response: under the same output budget, GD velocities scale with driving force, Muon drives all active modes at cap \(c\), and PhysMuon allocates the budget by channel-wise fluctuations (②). (d) Temporal memory: single relaxation time \(\tau\) in Muon versus parallel fast and slow relaxation modes \(M=wM^{f}+(1-w)M^{s}\) in PhysMuon (①).}
\label{fig:model}
\end{figure}

We treat a weight matrix \(W\in\mathbb R^{m\times n}\) (\(m\) output channels, \(n\) input channels) as a responsive medium mapping input \(x\) to output \(y=Wx\) (Fig.~\ref{fig:model}a). The scalar \(\Phi\) measuring the output-target mismatch is called the mismatch potential: for a linear output with squared mismatch, each output channel acts as a spring extended by its mismatch, and \(\Phi\) is the total spring energy; for a general neural network, \(\Phi\) is the current mini-batch loss. This potential exerts a driving force on each matrix entry,
\begin{equation}
X=-\frac{\partial\Phi}{\partial W},
\label{eq:force}
\end{equation}
which is the negative gradient. When the matrix rearranges at velocity \(V=\dot W\), the mismatch potential decreases at the dissipation rate \(P=\sum_{ij}X_{ij}V_{ij}\). The driving force \(X\) specifies only which rearrangements lower mismatch; how \(X\) becomes the rearrangement velocity \(V\) depends on the medium's spatial and temporal response.

Consider first the spatial rearrangement response. Because downstream layers observe only changes in this matrix's output, we constrain rearrangement velocity at the output: measuring signal amplitudes by the root-mean-square across channels, rearrangement must not change the output by more than \(\varepsilon\) times the input amplitude, a bound we call the output budget. By singular value decomposition \(X=\sum_k\sigma_ku_kv_k^{\mathsf T}\) (retaining only terms with \(\sigma_k>0\)), the driving force splits into uncoupled collective modes of coordinated matrix rearrangement in fixed proportions. The \(k\)th mode contributes the product of its driving force \(\sigma_k\) and velocity to the dissipation rate, while the output budget caps every mode's velocity at a common limit \(c\) set by \(\varepsilon\) and the matrix dimensions.

Consequently, the response that maximizes the dissipation rate drives every active mode at the velocity cap \(c\):
\begin{equation}
V^\star=c\sum_k u_kv_k^{\mathsf T}=c\,\mathrm{polar}(X),
\label{eq:polar}
\end{equation}
where \(\mathrm{polar}(X)\) is the orthogonal polar factor of \(X\), so equation~\eqref{eq:polar} recovers standard Muon's semi-orthogonalized update (see Supplementary Note~1 for the derivation). Here \(\sigma_k\) determines only how fast each mode releases mismatch potential, while the output budget sets all active mode velocities to \(c\). By contrast, ordinary gradient descent acts as a viscous medium whose velocity scales with driving force: under the same output budget only the strongest mode reaches \(c\), leaving weaker modes slow (Fig.~\ref{fig:model}c).

Consider next temporal stress accumulation and relaxation. In actual training, each step's driving force, called the instantaneous force, comes from a single mini-batch and fluctuates in direction from batch to batch, often changing before the matrix can rearrange. Matrix rearrangement is therefore governed by internal stress \(M\) accumulated from historical driving forces: components flipping sign across batches cancel, while components persisting in the same direction reinforce one another. Its simplest continuous evolution law is first-order relaxation \(\tau\dot M=X-M\) with relaxation time \(\tau\), classical Maxwell viscoelastic relaxation\cite{maxwell1867} (Fig.~\ref{fig:model}d).

Treating the driving force within each training step \(\Delta t\) as a constant \(X_t\), exact integration of this equation yields
\begin{equation}
M_t=\beta M_{t-1}+(1-\beta)X_t,\qquad \beta=e^{-\Delta t/\tau},
\label{eq:relax}
\end{equation}
which is momentum's exponential moving average of historical gradients\cite{polyak1964}, so we also refer to \(M\) as memory. Equation~\eqref{eq:polar} therefore acts on internal stress \(M_t\) rather than instantaneous force \(X_t\): instantaneous forces accumulate into stress before semi-orthogonalization, because reversing the order would amplify single-step noise directions to cap \(c\).

\subsection*{PhysMuon: fluctuation-calibrated output budget}

The two-step derivation above corresponds to the simplest idealized medium: spatially, the output budget treats all channels equally; temporally, internal stress has a single relaxation time \(\tau\). We first calibrate the spatial output budget to channel fluctuations (change ②). Because each step's driving force is averaged over a mini-batch, it carries sampling fluctuations that vary across channels. In an auxiliary model, we treat each matrix entry as an overdamped particle dragged by a fluctuating force: minimizing the upper bound on cumulative mismatch over a fixed interval sets each entry's optimal damping proportional to its force-fluctuation standard deviation. All else being equal, sharing a common damping across all entries, as in standard Muon's isotropic budget, is optimal only when fluctuations are uniform.

Under two simplifying assumptions, this optimal damping factorizes across the output and input sides (Supplementary Note~1); adopting these damping factors as quadratic output-budget weights transforms the maximum-dissipation response into
\begin{equation}
V^\star=c\,\bar S^{-1/4}\,\mathrm{polar}\!\big(\bar S^{-1/4}X\bar C^{-1/4}\big)\,\bar C^{-1/4},
\label{eq:iso}
\end{equation}
where \(\bar S\) and \(\bar C\) are mean-normalized diagonal matrices of mean squared output-channel backpropagated errors and input-channel activations, respectively. Splitting each step's samples into sub-batches, or micro-batches, lets us estimate both matrices directly from same-step micro-batch gradient differences; in multi-GPU training, per-GPU gradients serve directly as micro-batch gradients. Equation~\eqref{eq:iso} transforms into fluctuation-weighted coordinates for semi-orthogonalization and transforms back, slowing rearrangement in high-fluctuation channels and accelerating it in low-fluctuation ones.

\subsection*{PhysMuon: fast and slow relaxation modes}

\begin{figure}[!t]
\centering
\includegraphics[width=\textwidth]{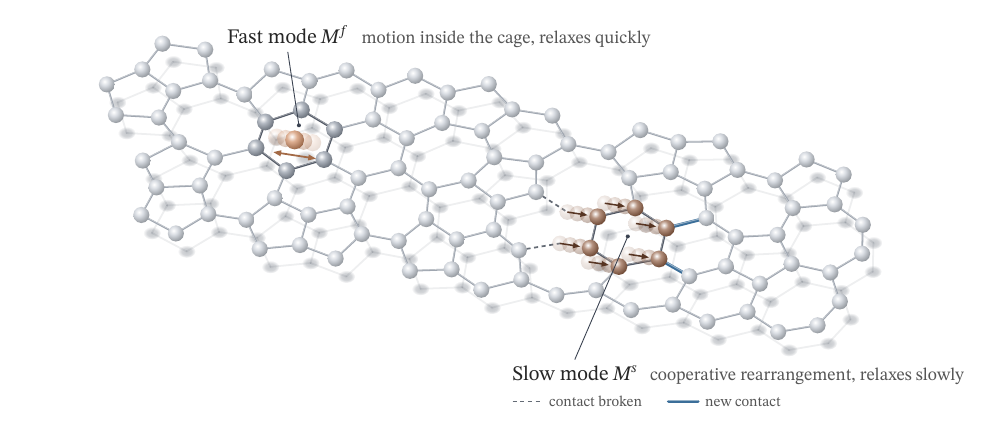}\\[2pt]
{\small (a)}\\[8pt]
\includegraphics[width=\textwidth]{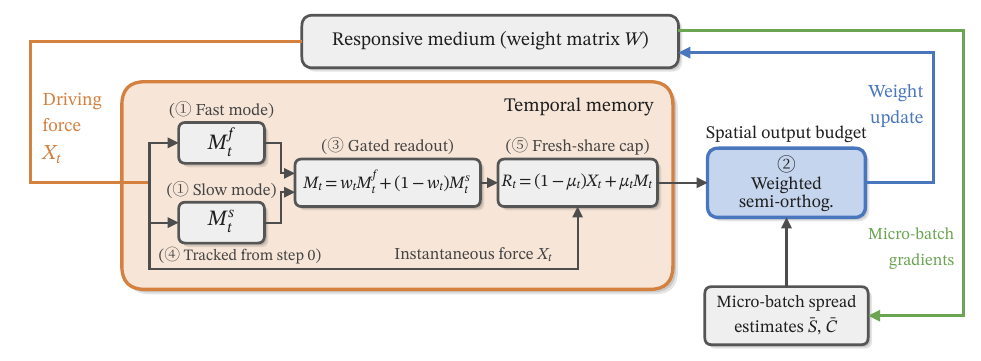}\\[2pt]
{\small (b)}
\caption[Physical picture and construction of PhysMuon]{Physical picture and single-step update of PhysMuon (schematic). (a) Analogy to fast and slow relaxation in glassy systems: fast localized motion within a nearest-neighbour cage versus slow cooperative rearrangement breaking existing contacts (dashed lines) and forming new ones (solid blue lines). Faint copies mark earlier positions and arrows show displacements. (b) Single-step PhysMuon update pipeline on weight matrix \(W\). The orange box shows temporal memory: the driving force feeds into fast and slow relaxation modes (①) tracked from step 0 (④), fast-mode weight \(w_t\) is set by rebound-to-drift strength (③), and readout memory mixes with instantaneous force into \(R_t\) under the fresh-share cap \(\varphi\) (⑤). The blue box shows the spatial budget: \(\bar S\) and \(\bar C\) are estimated from same-step micro-batch gradient differences, and semi-orthogonalization acts on \(R_t\) in weighted coordinates (②).}
\label{fig:physmuon}
\end{figure}

Temporally, standard Muon's internal stress has only a single relaxation time \(\tau\), coupling response speed to memory depth: small \(\tau\) tracks rapidly but forgets quickly, while large \(\tau\) remembers longer but responds sluggishly. In complex media, internal rearrangement typically involves multiple modes with distinct relaxation times, and the macroscopic stress is a weighted sum of mode stresses. In glassy systems, for example (Fig.~\ref{fig:physmuon}a), a single particle moves locally within its nearest-neighbour cage and relaxes rapidly, whereas multiple particles must break existing contacts before rearranging cooperatively and relax slowly\cite{berthier2011glass}. Motivated by this, change ① introduces one fast and one slow relaxation mode, the minimum needed to decouple response speed from memory depth:
\begin{equation}
M_t=w\,M^f_t+(1-w)\,M^s_t,
\label{eq:twomode}
\end{equation}
where fast mode \(M^f\) and slow mode \(M^s\) evolve independently via equation~\eqref{eq:relax} with \(\beta_f<\beta_s\), and \(w\) is the fast-mode readout weight; the fast mode handles short-timescale response while the slow mode maintains long-timescale memory (Supplementary Fig.~1).

Changes ③--⑤ govern memory readout. Beyond persistent drift across steps and batch-specific sampling noise, each step's instantaneous force contains a rebound component, an elastic restoring force that arises when the previous update overshoots local equilibrium and flips sign at every step. Memory evolving under equation~\eqref{eq:relax} transmits a fraction of this rebound, with smaller \(\beta\) leaking more: under language-model settings, the fast mode (\(\beta_f=0.85\)) transmits \(8.1\%\) of rebound amplitude, whereas the slow mode (\(\beta_s=0.985\)) transmits only \(0.76\%\). Reading more of the fast mode tracks drift faster but admits more rebound. To balance both, change ③ uses gated readout to measure rebound relative to drift for each matrix at every step and dynamically adjust \(w_t\) in equation~\eqref{eq:twomode}: stronger rebound yields smaller \(w_t\), with no explicit dependence on step count or learning rate.

Change ④ lets both modes track the driving force continuously from step \(0\), while switching on two-mode readout only at a preset onset step; before onset, updates still use standard single-mode momentum. The slow mode's state at onset thus reflects accumulation on its own timescale from step \(0\), rather than being copied from single-mode momentum.

Change ⑤ addresses an implementation detail of standard Muon: semi-orthogonalization receives not \(M_t\) alone, but a Nesterov-style mix \(R_t=(1-\mu)X_t\allowbreak+\mu M_t\) (\(\mu=0.95\) for language models), so instantaneous noise and rebound are only partly filtered. The more the slow mode is read out and the deeper the memory, the more thoroughly noise and rebound cancel in time averaging and the smaller \(\|M_t\|\) becomes. At fixed \(\mu\), the norm share of \(X_t\) in \(R_t\), called the fresh share, rises accordingly and lets more unfiltered rebound enter. Change ⑤ therefore raises \(\mu\) at the current step whenever needed to keep the fresh share at or below \(\varphi\).

Together, these five changes form PhysMuon (see Fig.~\ref{fig:model}b for trajectories on a mismatch potential and Fig.~\ref{fig:physmuon}b for the single-step pipeline). Each change generalizes one minimal assumption of standard Muon; disabling any change recovers the corresponding baseline form, and disabling all five recovers standard Muon. Constants across tasks are listed in Supplementary Table~3.

\subsection*{Language-model benchmark}

\begin{figure}[!t]
\centering
\includegraphics[width=\textwidth]{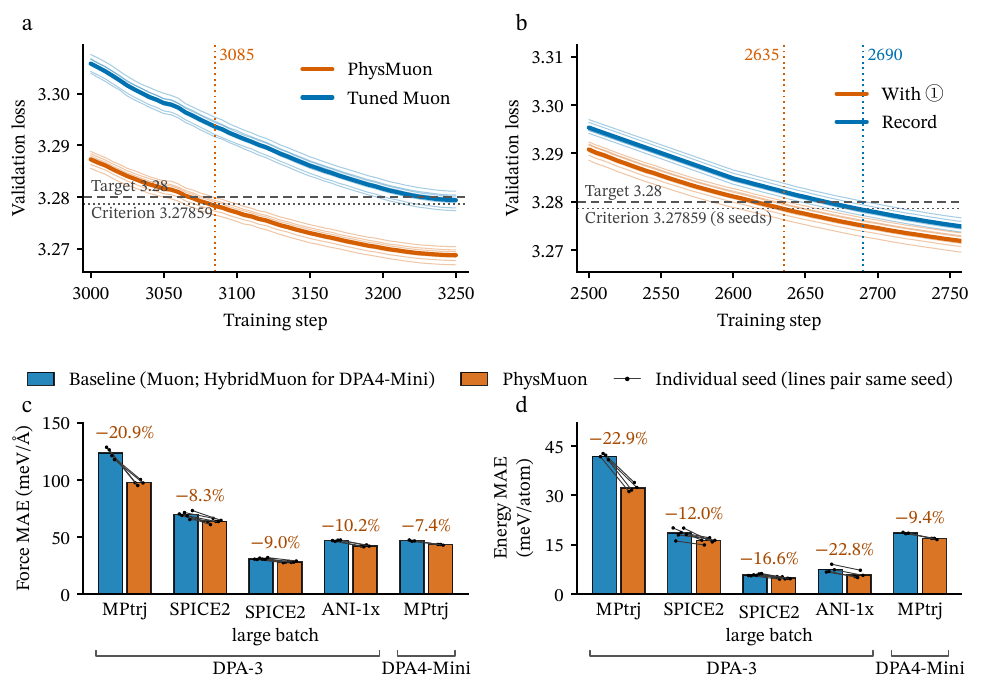}
\caption[Language model and interatomic potentials]{Training results on language models and machine-learned interatomic potentials. (a,b) GPT-2 124M training-speed benchmark (modded-nanogpt): (a) tuned Muon baseline versus full PhysMuon; (b) leaderboard record configuration versus replacing only momentum with fast and slow relaxation modes (①), evaluated with tail weight averaging. Thin lines show individual seeds and thick lines show seed averages; dashed, horizontal dotted, and vertical dotted lines mark the \(3.28\) target loss, significance threshold, and first-crossing steps, respectively. (c) Validation force MAE across five interatomic-potential tasks. Bars show seed means for each task: blue denotes the baseline, which is momentum-tuned Muon for DPA-3 and HybridMuon for DPA4-Mini, and orange denotes PhysMuon; dots mark individual seeds, thin lines connect paired runs, and numbers above bars show mean seed-paired relative changes. (d) Validation per-atom energy MAE for the same tasks.}
\label{fig:lm}
\end{figure}

We first evaluate PhysMuon on the public language-model training-speed benchmark modded-nanogpt~\cite{moddednanogpt}. This benchmark fixes the 124M-parameter GPT-2 architecture, training data, and tokens per step, comparing how many steps are needed to lower held-out validation loss to \(3.28\). Each configuration is trained across 8 random seeds; under the significance criterion in Methods, the first validation step at which seed-averaged loss meets the threshold is termed the first-crossing step, where lower values are better. With all five changes added to the tuned Muon baseline, PhysMuon reaches the target at step \(3085\), whereas the baseline does not reach it within \(3250\) steps (Fig.~\ref{fig:lm}a).

Using change ① to replace only the memory kernel also advances target attainment in the stronger leaderboard record configuration. At submission, the record configuration stacked SOAP~\cite{vyas2024soap} and other components on Muon, reaching a first-crossing step of \(2690\). Keeping all other components unchanged and replacing only momentum with fast and slow relaxation modes advances the first-crossing step by \(55\) steps to \(2635\) (Fig.~\ref{fig:lm}b); repeating the run independently on another GPU yields \(2645\).

On the 124M model, PhysMuon achieves a mean validation loss \(0.011\) lower than the Muon baseline at step \(3250\). Scaling to 302M parameters and training for \(3250\) steps widens this gap to \(0.018\), and PhysMuon achieves a lower validation loss in every replicate run.

\subsection*{Interatomic potentials}

Machine-learned interatomic potentials predict system energy and atomic forces from atomic configurations and form the basis of molecular dynamics simulations~\cite{zhang2018deepmd,batzner2022nequip,unke2021mlff,tokita2023train,cheng2024cace,xia2025evolution}. Using PhysMuon in deepmd-kit~\cite{zeng2025deepmdv3}, we completed five tasks: DPA-3~\cite{zhang2026dpa3} on MPtrj~\cite{deng2023chgnet}, regular-batch and large-batch SPICE2~\cite{eastman2024spice2,kovacs2025maceoff}, and ANI-1x~\cite{smith2020ani1x}, plus DPA4-Mini on MPtrj. The baseline is momentum-tuned Muon for DPA-3 and built-in HybridMuon for DPA4-Mini; within each task, both optimizers use identical training steps and random seeds. Task settings are detailed in Supplementary Note~3.

Across all replicate runs in all five tasks, PhysMuon lowers both validation force and per-atom energy errors in every run, with task-averaged reductions of \(7.4\%\)--\(20.9\%\) for forces and \(9.4\%\)--\(22.9\%\) for energies (Fig.~\ref{fig:lm}c,d and Table~\ref{tab:mlip}; per-seed values are in Supplementary Note~4).

\subsection*{Matbench Discovery}

Matbench Discovery~\cite{riebesell2025matbench} is a primary evaluation suite for universal interatomic potentials~\cite{takamoto2022pfp,chen2022m3gnet,gong2023examining,park2024sevennet,bochkarev2024grace,batatia2025macemp,mazitov2025petmad,barrosoluque2026omat24,yang2024mattersim,yin2025alphanet,kim2026crossdomain}; we test PhysMuon here with larger-scale training. The benchmark relaxes WBM candidate crystals to evaluate stability classification F1 from relaxed energies and structural agreement with density functional theory (DFT) relaxations via the root-mean-square-deviation area under the curve \(\mathrm{RMSD}_{\mathrm{AUC}}\), and then computes lattice thermal conductivity error \(\kappa_{\mathrm{SRME}}\)~\cite{pota2024thermal} across 103 crystals. The combined performance score CPS is a weighted combination of these three metrics; higher F1, \(\mathrm{RMSD}_{\mathrm{AUC}}\), and CPS are better, whereas lower \(\kappa_{\mathrm{SRME}}\) is better.

\begin{figure}[tp]
\centering
\includegraphics[width=\textwidth]{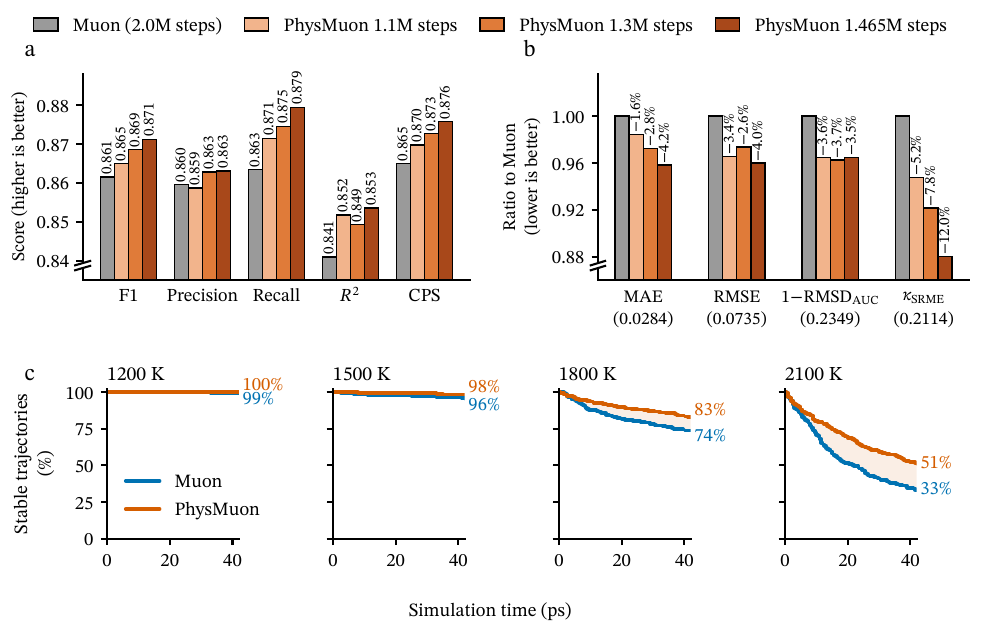}
\caption[Matbench Discovery and molecular dynamics]{Performance on Matbench Discovery and in high-temperature molecular dynamics. (a,b) DPA4-Pro metrics on Matbench Discovery. Grey bars show the official HybridMuon leaderboard entry at 2.0M steps and orange bars show PhysMuon exponential-moving-average checkpoints at 1.1M, 1.3M, and 1.465M steps, with both models trained on MPtrj alone. (a) Higher-is-better raw metrics (F1, precision, recall, \(R^2\), and CPS). (b) Lower-is-better metrics (formation-energy MAE and RMSE in eV per atom, \(1-\mathrm{RMSD}_{\mathrm{AUC}}\), and \(\kappa_{\mathrm{SRME}}\)), shown relative to Muon with Muon's raw values in parentheses. (c) Fraction of SPICE2 trajectories remaining stable over time at \(1200\)--\(2100\,\mathrm K\). Blue and orange curves show Muon and PhysMuon, respectively, across \(336\) trajectories per temperature and optimizer; end labels give stable percentages at \(42\,\mathrm{ps}\).}
\label{fig:matbench}
\end{figure}

A DPA4-Pro model trained with PhysMuon for 1.465 million steps on MPtrj alone achieves an F1 of \(0.8712\), \(\kappa_{\mathrm{SRME}}\) of \(0.1860\), \(\mathrm{RMSD}_{\mathrm{AUC}}\) of \(0.7734\), and CPS of \(0.8757\). By comparison, the official leaderboard DPA4-Pro entry~\cite{li2026dpa4} trained with HybridMuon for 2 million steps achieves \(0.8615\), \(0.2114\), \(0.7651\), and \(0.8650\), respectively (Fig.~\ref{fig:matbench}a,b). PhysMuon outperforms the official model on all three metrics and on CPS, exceeding every previously reported MPtrj-compliant model on Matbench Discovery in each metric; Supplementary Table~9 lists the top leaderboard entries. Even with fewer training steps, PhysMuon matches or exceeds Muon: CPS reaches \(0.8654\) at step 900k, and every fully evaluated checkpoint thereafter is at or above the official 2-million-step score of \(0.8650\), reaching \(0.8698\) by step 1.1M (Fig.~\ref{fig:matbench}a).

Of the \(0.0107\) gain in CPS, F1 and \(\kappa_{\mathrm{SRME}}\) each contribute nearly half. F1 is evaluated on candidate crystals outside the training set, while \(\kappa_{\mathrm{SRME}}\) depends on second and third energy derivatives with respect to atomic positions, which are not directly fitted during training~\cite{deng2025softening,loew2025phonons,dong2024gpumd}. Improvements on both metrics indicate that models trained with PhysMuon generalize better. Training and evaluation details are given in Methods.

\begin{table}[tp]
\centering
\caption[Errors of interatomic potentials]{Errors of machine-learned interatomic potentials. `Force MAE' lists baseline \(\to\) PhysMuon force mean absolute error (MAE) as mean \(\pm\) s.d. across seeds (baseline is HybridMuon for DPA4-Mini and Muon otherwise); `Force' and `Energy' give relative changes in force and per-atom energy MAE, and `Improved' gives the number of seeds with lower force error. The first five rows report seed-paired validation means; the `ANI-1x MD snapshots' row reports relative changes computed from averages over 360 snapshots across 4 model pairs against DFT references, with energy error evaluated after subtracting per-molecule constant offsets.}
\label{tab:mlip}
\footnotesize
\setlength{\tabcolsep}{3.5pt}
\begin{tabular}{@{}lcccc@{}}
\toprule
Task (model, training length) & Force MAE (meV/\AA) & Force & Improved & Energy \\
\midrule
MPtrj (DPA-3, 20,000 steps) & \(123.5\pm4.9 \to 97.6\pm2.1\) & $-20.9\%$ & 4/4 & $-22.9\%$ \\
SPICE2 (DPA-3, 8,000 steps) & \(69.6\pm2.7 \to 63.7\pm1.8\) & $-8.3\%$ & 6/6 & $-12.0\%$ \\
SPICE2 large batch (DPA-3, 20,000 steps) & \(30.9\pm0.7 \to 28.1\pm0.6\) & $-9.0\%$ & 6/6 & $-16.6\%$ \\
ANI-1x (DPA-3, 20,000 steps) & \(46.9\pm0.5 \to 42.2\pm0.7\) & $-10.2\%$ & 4/4 & $-22.8\%$ \\
MPtrj (DPA4-Mini, 10 epochs) & \(46.7\pm0.6 \to 43.3\pm0.4\) & $-7.4\%$ & 3/3 & $-9.4\%$ \\
\midrule
ANI-1x MD snapshots (360 snapshots) & \(38.0\pm0.8 \to 34.2\pm0.7\) & $-9.9\%$ & 4/4 & $-5.1\%$ \\
\bottomrule
\end{tabular}
\end{table}

\subsection*{Molecular dynamics: accuracy and stability}

Trained interatomic potentials are ultimately deployed in MD simulations~\cite{fu2023forces,morrow2023validate,musaelian2023allegro,frank2024euclidean,fu2025esen,bigi2025darkside,liu2023discrepancies}. We ran MD simulations on 20 molecules using all trained models from the ANI-1x and SPICE2 large-batch tasks, with the setup in Methods. To test accuracy on MD configurations, \(360\) configurations (\(120\) each at \(300\), \(1000\), and \(1500\,\mathrm K\)) were sampled from trajectories of the ANI-1x models and re-evaluated with DFT at the dataset's native \(\omega\)B97x/6-31G(d) level of theory. PhysMuon models lower mean force error by \(9.9\%\) relative to Muon models, from \(38.0\) to \(34.2\,\text{meV/\AA}\), and every PhysMuon model outperforms every Muon model (Table~\ref{tab:mlip}). Reductions across the three temperatures span \(9.7\%\)--\(10.3\%\), matching the \(10.2\%\) validation reduction.

PhysMuon models are also more stable. In regular tests across \(600\)--\(1500\,\mathrm K\), where \(576\) constant-temperature trajectories were run per optimizer, PhysMuon models incur zero instabilities whereas Muon models incur one unstable trajectory (Table~\ref{tab:md}). When the thermostat is turned off at \(300\,\mathrm K\) for \(40\,\mathrm{ps}\), both model families keep mean total-energy drifts at or below \(5\times10^{-4}\,\mathrm{meV}\) per atom per picosecond.

To compare both optimizers at higher temperatures, we ran stress tests on the SPICE2 models at \(1200\), \(1500\), \(1800\), and \(2100\,\mathrm K\)~\cite{stocker2022robust,yang2024reactive,chiang2025mlip}, with \(336\) trajectories of \(42\,\mathrm{ps}\) per temperature and optimizer. At every temperature, PhysMuon models incur fewer unstable trajectories than Muon models (Fig.~\ref{fig:matbench}c and Table~\ref{tab:md}). Unstable fractions fall from \(4.5\%\) to \(1.8\%\) at \(1500\,\mathrm K\), from \(26.2\%\) to \(17.3\%\) at \(1800\,\mathrm K\), and from \(66.7\%\) to \(48.8\%\) at \(2100\,\mathrm K\). Across all four temperatures combined, total unstable trajectories drop by \(30\%\), from \(329\) to \(229\). Evaluated over the 6 independently trained seeds, PhysMuon reduces instability in 5 of 6 seeds with 1 tie at \(1500\,\mathrm K\) and in 5 of 6 seeds at \(2100\,\mathrm K\). The corresponding instability rates fall by factors of \(2.5\) at \(1500\,\mathrm K\) and \(1.7\) at \(2100\,\mathrm K\). At \(2100\,\mathrm K\), PhysMuon lowers instabilities across all 8 test molecules and extends the 42-ps restricted mean survival time from \(23.4\) to \(29.4\,\mathrm{ps}\) (Supplementary Note~5). Linearly interpolating instability fractions shows that PhysMuon reaches Muon's \(1500\) and \(1800\,\mathrm K\) levels only at \(1552\) and \(1885\,\mathrm K\), an increase of \(50\)--\(85\,\mathrm K\) in tolerated temperature.

Direct DFT bond-stretching scans of Si--Cl, C--H, and Si--C, three bonds frequently broken under thermal stress, trace this stability gain to reduced energy softening far from equilibrium. Near equilibrium, both model families match DFT within \(0.05\,\mathrm{eV}\). At \(1\)--\(3\,\text{\AA}\) extensions, where potentials soften relative to DFT, PhysMuon models predict \(5.7\%\)--\(6.9\%\) higher stretching energies than Muon models and lower full-curve energy MAE by \(8.6\%\)--\(16.4\%\). This barrier lift matches the \(6.5\%\) increase in the effective Arrhenius dissociation barrier fitted across all SPICE2 stress-test instabilities (Supplementary Note~5).

\begin{table}[tp]
\centering
\caption[Unstable trajectories in molecular dynamics]{Unstable trajectories in molecular dynamics. `Trajectories' lists the number of test trajectories per optimizer; seed-level statistical comparisons and confidence intervals appear in Supplementary Note~5. `42-ps RMST' is the 42-ps restricted mean survival time, with trajectories remaining stable throughout the run capped at \(42\,\mathrm{ps}\).}
\label{tab:md}
\footnotesize
\setlength{\tabcolsep}{4.5pt}
\begin{tabular}{@{}llrcccc@{}}
\toprule
 & & & \multicolumn{2}{c}{Unstable trajectories} & \multicolumn{2}{c}{42-ps RMST (ps)} \\
\cmidrule(lr){4-5}\cmidrule(lr){6-7}
Test & Temperature & Trajectories & Muon & PhysMuon & Muon & PhysMuon \\
\midrule
ANI-1x, regular & \(300\,\mathrm K\) (NVE) & 48 & 0 & 0 & -- & -- \\
ANI-1x, regular & \(1000\,\mathrm K\) & 144 & 0 & 0 & -- & -- \\
ANI-1x, regular & \(1500\,\mathrm K\) & 144 & 1 (0.7\%) & 0 & -- & -- \\
SPICE2, regular & \(300\,\mathrm K\) (NVE) & 48 & 0 & 0 & -- & -- \\
SPICE2, regular & \(600\,\mathrm K\) & 144 & 0 & 0 & -- & -- \\
SPICE2, regular & \(1000\,\mathrm K\) & 144 & 0 & 0 & -- & -- \\
\midrule
SPICE2, stress & \(1200\,\mathrm K\) & 336 & 2 (0.6\%) & 1 (0.3\%) & 41.9 & 41.9 \\
SPICE2, stress & \(1500\,\mathrm K\) & 336 & 15 (4.5\%) & 6 (1.8\%) & 41.0 & 41.6 \\
SPICE2, stress & \(1800\,\mathrm K\) & 336 & 88 (26.2\%) & 58 (17.3\%) & 35.0 & 37.8 \\
SPICE2, stress & \(2100\,\mathrm K\) & 336 & 224 (66.7\%) & 164 (48.8\%) & 23.4 & 29.4 \\
\bottomrule
\end{tabular}
\end{table}
\section*{Discussion}\label{pm:discussion}

By viewing each weight matrix during training as a responsive medium with memory, we derived both components of Muon: semi-orthogonalization gives the response direction that maximizes dissipation rate under an output budget, while momentum is an internal stress with a single relaxation time. Replacing these two minimal forms with ones closer to a real medium yields PhysMuon. The output budget is calibrated to fluctuations (②); memory combines fast and slow relaxation modes that track the driving force from step 0 (①④) and is read out according to the relative strength of rebound versus drift (③), with a cap on the instantaneous-force share (⑤). Across language models, five interatomic-potential tasks, Matbench Discovery, and high-temperature molecular dynamics, PhysMuon consistently outperforms the baselines in steps to target, force and energy errors, materials discovery metrics, and simulation stability. These results show that optimizers can be derived and improved from the properties of a responsive medium.

The interatomic-potential and molecular-dynamics results show that PhysMuon improves accuracy both inside and outside the training distribution. On the SPICE2 large-batch task, validation force error drops from \(30.93\) to \(28.14\,\text{meV/\AA}\), and errors across all 12 elements fall by \(8.1\%\)--\(16.8\%\) (Supplementary Note~4). Under high-temperature stress and rigid bond stretching, the same two-mode configuration reduces energy softening at large bond extensions and raises both survival time and effective dissociation barriers. PhysMuon thus improves accuracy on configurations covered by the training data and extrapolation to far-from-equilibrium configurations where training data are sparse.

We also used a read-only memory probe that measures without altering training to track a proxy for the optimal memory duration of momentum, that is, how many recent gradient steps momentum should average. Across 8 independent language-model runs under the standard decaying learning-rate schedule, this proxy grows continuously during training, and in diagnostic runs growth persists after subtracting early-training rebound caused by the optimizer itself (Supplementary Note~6). Because shrinking learning rates and intrinsic landscape slowing both contribute, we treat this probe as a qualitative diagnostic of the dynamical slowing, which resembles physical aging in soft matter, where glasses and polymers relax more slowly the longer they wait after a quench~\cite{struik1978,bouchaud1992aging}.

These gains come with modest memory overhead. On GPT-2 124M, tracking two relaxation modes and fluctuation metrics adds \(0.65\,\mathrm{GiB}\) of persistent memory and increases per-step time by \(1.5\%\)--\(2.9\%\). In timing runs on the RTX 5090, all eight PhysMuon runs and three of four Muon runs reached the target validation loss within 3,250 steps. Among runs that reached the target, the mean time was \(8{,}228\,\mathrm{s}\) for PhysMuon and \(8{,}485\,\mathrm{s}\) for Muon. On DPA4-Pro, PhysMuon adds roughly \(175\,\mathrm{MiB}\) per GPU (Supplementary Note~7). Configuration transfers cleanly across interatomic-potential tasks: because gradient rebound is weaker than in language models, all six share the same two-mode gated readout without change ⑤. Five of the six tasks also share identical relaxation constants, with spike protection added for heavy-tailed crystal gradients. Selecting memory depths automatically across architectures is a direction for future work.

\section*{Methods}\label{pm:methods}

\subsection*{Language-model benchmark}
Language-model experiments follow the modded-nanogpt benchmark~\cite{moddednanogpt}. The model is a 124M-parameter GPT-2~\cite{radford2019gpt2} (12 layers, width 768, vocabulary size 50,304) trained on FineWeb~\cite{penedo2024fineweb}. Each optimization step processes 524,288 tokens in one forward and backward pass, with a target held-out validation loss of \(3.28\). Target attainment requires \((3.28-\bar L)\sqrt n\ge0.004\), where \(\bar L\) is the mean validation loss across \(n\) seeds at the same step. Under benchmark rules, the seed-to-seed spread at attainment is approximately \(0.0013\), so this threshold corresponds to a one-sided \(p\approx0.001\). The first-crossing step is the first validation step meeting this criterion. Validation runs every 5 steps near the target. Both the criterion and the validation grid were fixed before reading out formal runs; all seeds of a configuration share the same grid, and configurations are compared strictly at identical steps without interpolation.

Of the two benchmark configurations, the tuned Muon baseline is the official benchmark baseline (entry 36 in the results table), and the record configuration is the leaderboard record at submission (PR \#328). Their components and the constants used when replacing only the memory kernel are detailed in Supplementary Note~2. In Fig.~\ref{fig:lm}a, both groups used 8 seeds on a single RTX 5090, where Muon was rerun from the official baseline on that GPU. In Fig.~\ref{fig:lm}b, both groups used 8 seeds: the ①-only configuration ran on an A800, whereas the record came from PR \#328 submission logs on an A40. For the 302M model comparison, each configuration was trained across 3 seeds.

\subsection*{Implementation of PhysMuon}
PhysMuon acts separately on each two-dimensional hidden-layer weight matrix. Learning rates, weight decay, schedules, and all other parameter updates, such as AdamW in language models, remain identical to the replaced baseline. The implementation accumulates gradients \(G_t=-X_t\) and produces weight updates identical to those under the driving-force convention. The exponential moving averages are linear, the polar factor is an odd mapping, and \(w_t\), \(\mu_t\), \(\bar S\), and \(\bar C\) are invariant when gradients and memory flip sign together.

For temporal memory, the fast and slow modes \(M^f_t, M^s_t\) are updated from step 0 according to equation~\eqref{eq:relax} (④). Before the onset step \(T_{\mathrm{on}}\), updates use single-mode momentum \(M_t=\mu M_{t-1}+(1-\mu)G_t\); from \(T_{\mathrm{on}}\) onward, they switch to the two-mode memory \(M^{\mathrm{eff}}_t=w_tM^f_t+(1-w_t)M^s_t\) (①). Gated readout (③) sets the fast-mode weight to
\begin{equation}
w_t=w_{\max}\Big(1-\rho_f\,\frac{o_t}{d_t}\Big)_+,\qquad \rho_f=\frac{1-\beta_f}{1+\beta_f},
\label{eq:gate}
\end{equation}
where \((z)_+=\max(z,0)\), \(o_t=\big(-\cos(G_t,M^f_{t-1})\big)_+\) measures anti-alignment between the current gradient and the fast mode, and \(d_t=\|M^s_{t-1}\|/\|G_t\|\) is the norm ratio of the slow mode to the current gradient. Both quantities are computed step by step on each matrix and smoothed by an exponential moving average with decay \(0.9\). Change ⑤ first advances the fast and slow modes and then sets
\begin{equation}
\mu_t=\max\!\Big(\mu,\ \frac{1}{1+\frac{\varphi}{1-\varphi}\,\|M^{\mathrm{eff}}_t\|/\|G_t\|}\Big),
\label{eq:fresh}
\end{equation}
forming \(R_t=(1-\mu_t)G_t+\mu_tM^{\mathrm{eff}}_t\) so that the norm share of the instantaneous gradient in \(R_t\) does not exceed \(\varphi\).

For the spatial budget (②), each step's samples are split into \(K\) micro-batches without changing the summed gradient. For each hidden matrix, the row and column means of the between-micro-batch gradient variance estimate \(S_\alpha\) and \(C_i\), respectively, up to a channel-independent common factor and are smoothed by an exponential moving average with decay \(0.95\). After mean normalization, the diagonal entries of the weighting metrics \(\bar S^{1/2}\) and \(\bar C^{1/2}\) are clipped to \([1/2,2]\), while their \(-1/2\) powers appear in equation~\eqref{eq:iso}. Polar decomposition follows Muon's Newton--Schulz iteration~\cite{amsel2025polarexpress}. Because the theory specifies only the update direction, the output obtained by substituting \(R_t\) into equation~\eqref{eq:iso} is normalized to a Frobenius norm of \(\sqrt{\min(m,n)}\) and multiplied by Muon's shape factor \(\max(1,m/n)^{1/2}\), so the overall step scale matches that of Muon. Task constants are listed in Supplementary Table~3, and computational overhead is reported in Supplementary Note~7.

\subsection*{Interatomic potentials}
All five interatomic-potential tasks ran in deepmd-kit, where only the two-dimensional weight-matrix updates in each task's native Muon-family optimizer were replaced with PhysMuon (Supplementary Note~3). Because gradient rebound is weaker on these tasks than in language models, the fresh-share cap ⑤ was omitted; all five benchmark tasks and DPA4-Pro used two-mode gated readout with changes ①--④ and \(w_{\max}=0.3\). Regular-batch SPICE2 was trained for 8,000 steps, whereas MPtrj, large-batch SPICE2, and ANI-1x were trained for 20,000 steps; within each task, both optimizers ran on the same GPU model. The learning rate decayed exponentially from \(10^{-3}\) to \(3\times10^{-5}\), both scaled by a task-specific multiplier. Batch sizes were set in \texttt{auto} mode, which selects configurations per batch until the total atom count reaches at least a given threshold: regular-batch SPICE2 used \texttt{auto:256} with multiplier 16, MPtrj used \texttt{auto:1024} with multiplier 8, and large-batch SPICE2 and ANI-1x used \texttt{auto:2048} with multiplier 8.

DPA4-Mini used the official 10-epoch configuration (69,740 steps, 2 GPUs, gradient clipping threshold 5). Validation force and energy errors were evaluated with \texttt{dp test}, using exponential moving average weights for DPA4-Mini. Dataset versions and material- or molecule-level validation splits are detailed in Supplementary Note~3 and Supplementary Table~4.

\subsection*{Molecular dynamics tests}
Models came from the final checkpoints of each seed in the ANI-1x and SPICE2 large-batch tasks. Test molecules were selected from the respective validation sets (Supplementary Note~5): 12 for ANI-1x (10--27 atoms) and 8 for SPICE2 (18--60 atoms). Every trajectory ran for \(42\,\mathrm{ps}\) with a \(0.5\,\mathrm{fs}\) timestep, using a Langevin thermostat (BAOAB scheme, friction coefficient \(0.01\,\mathrm{fs^{-1}}\)) during constant-temperature segments. For each molecule, all models shared identical initial velocities and random-force sequences.

In regular tests, 7 trajectories were run per model and molecule: 1 trajectory equilibrated at \(300\,\mathrm K\) for \(2\,\mathrm{ps}\) followed by \(40\,\mathrm{ps}\) with the thermostat off, and 3 constant-temperature trajectories at each of two temperatures, \(1000\) and \(1500\,\mathrm K\) for ANI-1x and \(600\) and \(1000\,\mathrm K\) for SPICE2. In stress tests on SPICE2, all 7 trajectories per model and molecule ran at a single constant temperature, testing \(1200\), \(1500\), \(1800\), and \(2100\,\mathrm K\) separately.

A trajectory was classified as unstable if the molecule dissociated into two or more fragments (bonding threshold set to 1.5 times the sum of covalent radii), if any interatomic distance fell below \(0.6\,\text{\AA}\), or if the energy became non-finite, such as infinite or NaN. In stress tests, because thermal fluctuations can briefly stretch bonds past the bonding threshold, dissociation had to persist for at least \(100\,\mathrm{fs}\) to be classified as unstable. MD snapshots for accuracy evaluation were sampled from regular-test trajectories of seed-0 Muon and PhysMuon models on ANI-1x. We took 1 trajectory per temperature, using the thermostat-off run at \(300\,\mathrm K\); after equilibration, we sampled one configuration every \(8\,\mathrm{ps}\) and computed reference DFT energies and forces with PySCF~\cite{sun2020pyscf} (Supplementary Note~5).

\subsection*{Matbench Discovery}
Our DPA4-Pro model uses the DPA-4.0.1-Pro architecture~\cite{li2026dpa4} and was trained from scratch on MPtrj~\cite{deng2023chgnet} alone using the official schedule, loss weights, gradient clipping, and weight exponential moving average. We held out 1\% of materials by material identifier mp-id for validation (1,459 materials, 16,051 frames) and trained on the remaining 1,564,344 frames with PhysMuon. As in other tasks, PhysMuon was switched on at 30.77\% of the 2-million total steps, with single-mode momentum \(0.95\) beforehand and ② enabled at onset. Global gradient norms were clipped to 5 before each optimizer update. On any step whose pre-clipping norm exceeded 10 times the running median or overflowed float32, the slow mode skipped the step while the fast mode updated on the clipped gradient. From step 1.42M onward, ②'s pre-clipping fluctuation estimate skipped the same steps. Across GPU layouts, synchronized gradient accumulation maintained the official 16-GPU effective global batch size. The main results were taken from step 1.465M (Supplementary Note~4 and Supplementary Table~8). The baseline is the official Matbench Discovery leaderboard result; evaluating the official public checkpoint with the same relaxation and phono3py scripts reproduces its leaderboard scores.

Evaluation followed the official Matbench Discovery workflow~\cite{riebesell2025matbench}. We relaxed all 256,963 WBM test structures with the model, determined stability from relaxed energies, computed F1 on the 215,488 unique structural prototypes by convention, and evaluated \(\mathrm{RMSD}_{\mathrm{AUC}}=\frac{1}{N}\sum_{i=1}^N \max(1-\mathrm{RMSD}_i/0.05,\,0)\) from the relaxed structures. We relaxed 103 crystals under the protocol published by the DPA-4 authors and computed lattice thermal conductivity at \(300\,\mathrm K\) with phono3py to obtain \(\kappa_{\mathrm{SRME}}\)~\cite{pota2024thermal}. The combined performance score is \(\mathrm{CPS}=0.5\,\mathrm{F1}+0.4\,(1-\kappa_{\mathrm{SRME}}/2)+0.1\,\mathrm{RMSD}_{\mathrm{AUC}}\).

\section*{Author Contributions}
H.Y. supervised the project. Y.H. conceived the physical model, designed and implemented PhysMuon, performed the experiments and analyses, and drafted the manuscript. H.X., X.G., and H.Y. contributed to discussions and manuscript preparation. All authors reviewed and approved the final manuscript.

\section*{Funding}
We acknowledge financial support from NSFC (grant no. 12188101), Shanghai Science and Technology Program (no. 23JC1400900), the Guangdong Major Project of the Basic and Applied Basic Research (Future functional materials under extreme conditions, 2021B0301030005), Shanghai Pilot Program for Basic Research, Fudan University 21TQ1400100 (23TQ017), the robotic AI-Scientist platform of Chinese Academy of Science, and New Cornerstone Science Foundation.

\section*{Acknowledgements}
We thank Dr. Ruichen Jiang of Google Research in New York City for helpful discussions and assistance.

\section*{Data Availability}
Per-seed logs and summaries for the language-model and interatomic-potential experiments, along with molecular-dynamics trajectory statistics and DFT results, are publicly available at \url{https://github.com/Hyyu-Lab/PhysMuon}. Per-seed logs for the experiments using change ① to replace only the memory kernel on the record configuration and the tuned Muon baseline are also available in PRs \#339 and \#340 of the benchmark repository~\cite{moddednanogpt}, and logs for change ② alone are available in PR \#370. Pretrained DPA-4.0.1-Pro-MPtrj-Physmuon checkpoints for the 900k-step and 1.465M-step exponential moving averages are publicly available on Hugging Face at \url{https://huggingface.co/orange4664/DPA-4.0.1-Pro-MPtrj-Physmuon}.

\section*{Code Availability}
The PhysMuon implementation, deepmd-kit integration hooks, and reproduction scripts are publicly available at \url{https://github.com/Hyyu-Lab/PhysMuon}, and the training scripts submitted to the language-model benchmark are also available in PRs \#339, \#340, and \#370. Native PhysMuon support for deepmd-kit's \texttt{HybridMuon} optimizer has been submitted upstream at \url{https://github.com/deepmodeling/deepmd-kit/pull/6162}.

\section*{Competing Interests}
The authors declare no competing interests.

\FloatBarrier
\label{pm:refs}
\bibliographystyle{unsrtnat}
\bibliography{references}

\end{document}


\maketitle
\vspace{-10pt}
\tableofcontents
\clearpage

\section{Derivation of the physical model}\label{sn:model}

This note provides the derivations omitted from the main-text sections `Deriving Muon from a responsive medium' and the two PhysMuon subsections. Notation follows the main text: weight matrix \(W\in\mathbb R^{m\times n}\), driving force \(X=-\partial\Phi/\partial W\), rearrangement velocity \(V=\dot W\), and dissipation rate \(P=\sum_{ij}X_{ij}V_{ij}\).

\subsection{Driving force}

In the case of linear output and squared mismatch, the response of output channel \(\alpha\) is \(y_\alpha=\sum_iW_{\alpha i}x_i\) and the mismatch is \(e_\alpha=y_\alpha-y^\star_\alpha\); taking unit spring stiffness, the mismatch potential is the sum of spring potential energies across channels, \(\Phi=\tfrac12\sum_\alpha e_\alpha^2\). Perturbing matrix entry \(W_{\alpha i}\) by \(\delta W_{\alpha i}\) alters only the mismatch of channel \(\alpha\), \(\delta e_\alpha=x_i\,\delta W_{\alpha i}\), giving a first-order change in mismatch potential \(\delta\Phi=e_\alpha\,\delta e_\alpha=e_\alpha x_i\,\delta W_{\alpha i}\). The driving force on \(W_{\alpha i}\) is therefore \(X_{\alpha i}=-e_\alpha x_i\): the stronger the input and the larger the output mismatch, the stronger the driving force; in matrix form, \(X=-e\,x^{\mathsf T}\).

For an arbitrary differentiable network trained on a mini-batch of \(B\) samples, let a linear block map the \(b\)th sample as \(y_b=Wx_b\), with per-sample loss \(\ell_b\) and error backpropagated to the block's output \(\delta_b=\partial\ell_b/\partial y_b\). Taking the mismatch potential \(\Phi\) to be the mini-batch mean loss, the same definition yields
\begin{equation}
X=-\frac{\partial\Phi}{\partial W}=-\frac1B\sum_{b=1}^B\delta_b\,x_b^{\mathsf T},
\label{eq:force-general}
\end{equation}
which is the negative gradient of this block. For a single sample with linear output and squared mismatch, equation~\eqref{eq:force-general} reduces to \(X=-e\,x^{\mathsf T}\).

\subsection{Output budget and maximum-dissipation response}\label{ss:budget}

Because downstream layers sense only changes in the output of this block, the constraint on rearrangement velocity \(V=\dot W\) is placed at the output end. Measuring signal amplitudes by the root-mean-square across channels, we write the input amplitude as \(A_{\mathrm{in}}(x)=(\frac1n\sum_ix_i^2)^{1/2}\) and the output-change amplitude induced by rearrangement as \(A_{\mathrm{out}}(\dot y)=(\frac1m\sum_\alpha\dot y_\alpha^2)^{1/2}\) (where \(\dot y=Vx\)), and require for every input \(x\) that
\begin{equation}
A_{\mathrm{out}}(Vx)\le\varepsilon\,A_{\mathrm{in}}(x).
\label{eq:budget}
\end{equation}
For non-zero inputs, equation~\eqref{eq:budget} is equivalent to \(\|Vx\|/\|x\|\le c\), where
\begin{equation}
c\equiv\varepsilon\sqrt{m/n},
\label{eq:c}
\end{equation}
which requires the spectral norm of \(V\) not to exceed \(c\). This constant \(c\) is the output-budget cap in the main text, determined jointly by \(\varepsilon\) and the matrix dimensions.

The singular value decomposition of the driving force, \(X=\sum_{k=1}^r\sigma_ku_kv_k^{\mathsf T}\) (\(\sigma_k>0\), with orthonormal sets \(\{u_k\}\) and \(\{v_k\}\)), decomposes it into orthogonal collective modes: the \(k\)th mode \(u_kv_k^{\mathsf T}\) corresponds to a coordinated rearrangement of matrix entries \(W_{\alpha i}\) in proportion to \((u_k)_\alpha(v_k)_i\), into which only the component of input \(x\) along \(v_k\) enters, producing an output change along \(u_k\). Orthonormality of \(\{v_k\}\) on the input side prevents cross-talk among modes, while orthonormality of \(\{u_k\}\) on the output side eliminates cross-terms in the output-change amplitude.

For any \(V\) satisfying equation~\eqref{eq:budget}, we have \(u_k^{\mathsf T}Vv_k\le\|Vv_k\|\le c\), and therefore
\begin{equation}
P=\sum_{ij}X_{ij}V_{ij}=\sum_k\sigma_k\,u_k^{\mathsf T}Vv_k\le c\sum_k\sigma_k.
\label{eq:pbound}
\end{equation}
Setting \(V^\star=c\sum_ku_kv_k^{\mathsf T}=c\,\mathrm{polar}(X)\) gives a spectral norm of exactly \(c\), satisfying equation~\eqref{eq:budget} while achieving equality in equation~\eqref{eq:pbound}. Under the output budget, the response that maximizes the dissipation rate is thus given by main-text equation~(2), consistent with the characterization by Bernstein and Newhouse of the steepest-descent direction under the spectral norm~\cite{bernstein2024anthology}. Because the polar factor is an odd mapping, writing the discrete update \(W_{t+1}=W_t+\eta V^\star\) in terms of the gradient \(G=-X\) yields Muon's update \(W_{t+1}=W_t-\eta c\,\mathrm{polar}(G)\)~\cite{jordan2024muon}.

By contrast, in a viscous medium corresponding to ordinary gradient descent, rearrangement velocity is proportional to driving force, \(V=X/\gamma\); letting \(\sigma_1\) denote the largest singular value, equation~\eqref{eq:budget} then requires \(\sigma_1/\gamma\le c\). Taking \(\gamma=\sigma_1/c\) to saturate the output budget allows only the strongest mode to reach the velocity cap \(c\), while the \(k\)th mode moves at velocity \(c\,\sigma_k/\sigma_1\), yielding a total dissipation rate of \(c\sum_k\sigma_k^2/\sigma_1\). Because \(\sigma_k\le\sigma_1\), this dissipation rate never exceeds the upper bound in equation~\eqref{eq:pbound} and achieves equality if and only if all \(\sigma_k\) are equal (main-text Fig.~1c).

\subsection{Internal stress and momentum}

The simplest continuous evolution of internal stress \(M\) is first-order relaxation with relaxation time \(\tau>0\):
\begin{equation}
\tau\dot M=X(t)-M(t).
\label{eq:relax-cont}
\end{equation}
This evolution has three properties: when \(X\) remains constant over a long time, \(M\) approaches \(X\); when \(X\) fluctuates rapidly back and forth, \(M\) tracks only its slowly varying component while filtering out high-frequency fluctuations; if the driving force is removed at \(t=0\), stress decays as \(M(t)=M(0)\,e^{-t/\tau}\), so that larger \(\tau\) gives longer memory whereas smaller \(\tau\) gives faster response. Treating the driving force within each training step \(\Delta t\) as a constant \(X_t\), exact integration of equation~\eqref{eq:relax-cont} yields main-text equation~(3), the exponential moving average of momentum~\cite{polyak1964} with \(\beta=e^{-\Delta t/\tau}\); when \(\Delta t\ll\tau\), \(\beta\approx1-\Delta t/\tau\).

Because main-text equation~(2) drives all active modes to the velocity cap \(c\), the output budget should not be spent on transient single-step noise directions. Accumulating internal stress prior to semi-orthogonalization performs temporal filtering before spatial rearrangement: only driving-force components that persist in the same direction across multiple steps are retained, whereas random fluctuations cancel in the average. Conversely, if each step's instantaneous force were semi-orthogonalized before averaging, transient noise directions would already be amplified to the cap \(c\) before entering the average.

\subsection{Relaxation spectrum and two modes (①)}

Internal rearrangement in complex media typically involves multiple relaxation modes. Near a stable state, both the free energy and the dissipation function of internal degrees of freedom are positive-definite quadratic forms; decomposing into normal modes, each mode independently satisfies equation~\eqref{eq:relax-cont} with its own relaxation time \(\tau_j\): \(\tau_j\dot M_j=X-M_j\). The macroscopic stress is a superposition of modes with non-negative weights, normalized by the total static response:
\begin{equation}
M=\sum_jw_jM_j,\qquad w_j\ge0,\quad \sum_jw_j=1.
\label{eq:spectrum}
\end{equation}
In such a generalized Maxwell medium~\cite{maxwell1867}, non-negative weights ensure that every mode bears stress in the same direction under sustained loading, while weights summing to 1 ensure that \(M\) approaches \(X\) under a constant driving force applied for sufficiently long; when a constant driving force \(X_0\) is applied until equilibrium and removed at \(t=0\), \(M(t)=X_0\sum_jw_je^{-t/\tau_j}\), whose decay is characterized by a positive relaxation-time spectrum. Standard Muon retains only a single mode, tying response speed and memory depth to the same \(\tau\); decoupling the two requires at least one fast and one slow relaxation mode:
\begin{equation}
\tau_f\dot M^f=X-M^f,\qquad \tau_s\dot M^s=X-M^s,\qquad \tau_f<\tau_s,
\label{eq:twomode-cont}
\end{equation}
with combined stress \(M=wM^f+(1-w)M^s\) (\(0\le w\le 1\)), which is main-text equation~(5). Discretizing in the same manner as equation~\eqref{eq:relax-cont} gives decay coefficients \(\beta_f=e^{-\Delta t/\tau_f}\) and \(\beta_s=e^{-\Delta t/\tau_s}\); the response under application and removal of a constant driving force is illustrated in Supplementary Fig.~\ref{sfig:relax}.

\begin{figure}[tp]
\centering
\includegraphics[width=0.82\textwidth]{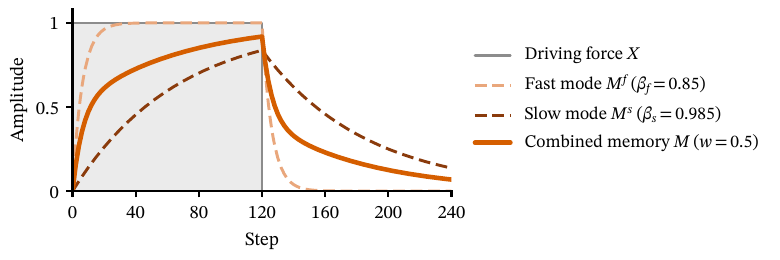}
\caption{Response of the fast and slow relaxation modes (schematic). A constant driving force (grey line, with light-grey shading indicating the loading interval) is applied at step 0 and removed at step 120; the fast mode (light-orange dashed line, decay coefficient \(0.85\), relaxation time \(\approx6\) steps), the slow mode (dark-orange dashed line, decay coefficient \(0.985\), relaxation time \(\approx66\) steps), and their combined memory with \(w=0.5\) (solid orange line) build up and relax accordingly: the combined memory rises rapidly with the fast mode at the onset of loading and maintains a long-tailed decay via the slow mode after the driving force is removed, balancing the response speed of the fast mode with the memory depth of the slow mode.}
\label{sfig:relax}
\end{figure}

\subsection{Fluctuation-calibrated output budget (②)}

\paragraph{Weighted output budget} Taking positive-definite diagonal matrices \(g=\mathrm{diag}(g_\alpha)\) on the output side and \(h=\mathrm{diag}(h_i)\) on the input side as weighting metrics, we generalize the signal amplitudes in equation~\eqref{eq:budget} to weighted amplitudes \(A^{(h)}_{\mathrm{in}}(x)=(\frac1n\sum_ix_i^2/h_i)^{1/2}\) and \(A^{(g)}_{\mathrm{out}}(\dot y)=(\frac1m\sum_\alpha g_\alpha\dot y_\alpha^2)^{1/2}\), and require for any input \(x\) that
\begin{equation}
A^{(g)}_{\mathrm{out}}(Vx)\le\varepsilon\,A^{(h)}_{\mathrm{in}}(x).
\label{eq:wbudget}
\end{equation}
When \(g=h=I\), equation~\eqref{eq:wbudget} reduces to equation~\eqref{eq:budget}. Introducing weighted coordinates \(\tilde x=h^{-1/2}x\) and \(\tilde y=g^{1/2}y\), the rearrangement velocity and driving force transform as \(\tilde V=g^{1/2}Vh^{1/2}\) and \(\tilde X=g^{-1/2}Xh^{-1/2}\): the dissipation rate is invariant (\(P=\sum_{\alpha i} X_{\alpha i}V_{\alpha i}=\sum_{\alpha i}\tilde X_{\alpha i}\tilde V_{\alpha i}\)), while equation~\eqref{eq:wbudget} becomes \(\|\tilde V\tilde x\|/\|\tilde x\|\le c\) in the new coordinates, identical in form to the spectral-norm constraint on \(V\) in equation~\eqref{eq:budget}. Applying the proof of Section~\ref{ss:budget} in weighted coordinates yields \(\tilde V^\star=c\,\mathrm{polar}(\tilde X)\), and transforming back to the original coordinates gives the maximum-dissipation response
\begin{equation}
V^\star=c\,g^{-1/2}\,\mathrm{polar}\!\big(g^{-1/2}Xh^{-1/2}\big)\,h^{-1/2}.
\label{eq:wpolar}
\end{equation}
Because the scalings inside and outside the polar factor arise from the same coordinate transformation, they necessarily share the same pair of \(g^{-1/2}\) and \(h^{-1/2}\).

\paragraph{Auxiliary model} To determine the metrics \(g\) and \(h\), consider an overdamped particle driven by a fluctuating force. Taking a quadratic approximation near a local minimum along the coordinate of a single matrix entry, let \(r\) denote the displacement of that entry from the local minimum (initial value \(r_0=\Delta\)), with mismatch potential \(\tfrac12kr^2\) (\(k>0\)) along that direction. Measuring time in optimization steps, with independent per-step force fluctuations of variance \(\sigma^2\), damping coefficient \(\gamma>0\), and \(B_t\) denoting standard Brownian motion, the continuous-time overdamped dynamics takes the form
\begin{equation}
dr_t=-\frac k\gamma\,r_t\,dt+\frac\sigma\gamma\,dB_t,\qquad r_0=\Delta.
\label{eq:aux}
\end{equation}
The second moment \(m(t)=\mathbb Er_t^2\) satisfies \(\dot m=-(2k/\gamma)m+\sigma^2/\gamma^2\); integrating over \([0,T]\) yields the cumulative-mismatch identity
\begin{equation}
J_T(\gamma)\equiv\mathbb E\int_0^T\tfrac12k\,r_t^2\,dt=\frac{\gamma\Delta^2}4+\frac{T\sigma^2}{4\gamma}-\frac\gamma4\,\mathbb Er_T^2,
\label{eq:jt}
\end{equation}
which drops the non-positive terminal term \(-\gamma\mathbb Er_T^2/4\) to yield the curvature-independent upper bound \(J_T(\gamma)\le\gamma\Delta^2/4+T\sigma^2/(4\gamma)\). This upper bound is minimized at
\begin{equation}
\gamma_0=\frac{\sigma\sqrt T}{|\Delta|},
\label{eq:gamma0}
\end{equation}
where the two terms are equal: the first term is proportional to \(\gamma\), reflecting slow release of initial mismatch under excessive damping, whereas the second term is inversely proportional to \(\gamma\), reflecting back-and-forth jitter driven by force fluctuations under insufficient damping. Thus, the damping that minimizes the upper bound on cumulative mismatch is proportional to the standard deviation \(\sigma\) of force fluctuations rather than the variance.

\paragraph{From matrix entries to channels} The variance \(\sigma_{\alpha i}^2\) of single-step mini-batch force fluctuations is proportional to the variance of the single-sample force component \(-\delta_\alpha x_i\), which satisfies the identity
\begin{equation}
\mathrm{Var}(\delta_\alpha x_i)=S_\alpha C_i+\mathrm{Cov}(\delta_\alpha^2,x_i^2)-\big(\mathbb E[\delta_\alpha x_i]\big)^2,
\label{eq:var}
\end{equation}
where \(S_\alpha=\mathbb E[\delta_\alpha^2]\) and \(C_i=\mathbb E[x_i^2]\) are the mean squared backpropagated error at output channel \(\alpha\) and the mean squared activation at input channel \(i\), respectively. We introduce two simplifying assumptions: Assumption A, that the last two terms in equation~\eqref{eq:var} are negligible compared with \(S_\alpha C_i\), which holds for example when backpropagated errors and input activations are approximately independent and the mean force is small; and Assumption B, that variations in initial displacement \(|\Delta_{\alpha i}|\) across matrix entries are negligible. Under these two assumptions, the optimal damping from equation~\eqref{eq:gamma0}, \(\gamma_{\alpha i}\propto\sqrt{S_\alpha C_i}=\sqrt{S_\alpha}\sqrt{C_i}\), factorizes across the output and input sides. As a modelling assumption, we take the damping coefficient of this auxiliary model as the quadratic-form weight in the output budget, under which each entry's velocity scales as the \(-1/2\) power of its weight under the budget constraint; setting the diagonal metrics on the two sides of equation~\eqref{eq:wbudget} to \(g=\bar S^{1/2}\) and \(h=\bar C^{1/2}\), where \(\bar S\) and \(\bar C\) are \(\mathrm{diag}(S_\alpha)\) and \(\mathrm{diag}(C_i)\) normalized by their channel means with the overall scale absorbed into \(c\) so that only relative magnitudes across channels matter, and substituting into equation~\eqref{eq:wpolar} yields main-text equation~(4). More generally, if \(|\Delta_{\alpha i}|\propto\sigma_{\alpha i}^\kappa\), then \(\gamma_{\alpha i}\propto\sigma_{\alpha i}^{1-\kappa}\), corresponding to metrics \(g=\bar S^\lambda\) and \(h=\bar C^\lambda\) (\(\lambda=(1-\kappa)/2\)): \(\kappa=0\) corresponds to Assumption B (\(\lambda=1/2\)), whereas \(\kappa=1\) gives \(\lambda=0\), recovering standard Muon.

\paragraph{Cost of non-uniform fluctuations} Consider \(N\) matrix entries sharing the same initial displacement magnitude \(|\Delta|\) and evolution duration \(T\), differing only in their force-fluctuation standard deviations \(\sigma_1,\dots,\sigma_N\). By equations~\eqref{eq:jt} and~\eqref{eq:gamma0}, when each entry takes its own optimal damping, the total cumulative-mismatch upper bound is \(\tfrac12|\Delta|\sqrt T\allowbreak\sum_{q=1}^N\sigma_q\); when all entries are forced to share a common damping as in the isotropic budget of standard Muon at the optimal common value, the total upper bound is \(\tfrac12|\Delta|\sqrt T\,(N\sum_{q=1}^N\sigma_q^2)^{1/2}\). Denoting the arithmetic mean across the \(N\) entries by \(\langle\cdot\rangle\), the Cauchy--Schwarz inequality gives the ratio of the latter to the former as \(\sqrt{\langle\sigma^2\rangle}/\langle\sigma\rangle\ge1\), with equality if and only if all \(\sigma_q\) are equal. An isotropic common damping is therefore optimal only when fluctuations are completely uniform; the more disparate the fluctuations, the larger the reduction in the cumulative-mismatch upper bound achieved by calibrating damping to fluctuations.

\paragraph{Combining with memory} With temporal memory included, the driving force substituted into main-text equation~(4) is replaced by the mix \(R_t\) of memory and instantaneous force (Section~\ref{ss:readout}). Under the quasi-static approximation that the smoothed metrics \(g\) and \(h\) vary slowly over the memory window, the spatial weighting map \(Z\mapsto g^{-1/2}Zh^{-1/2}\) is linear in the matrix at each step: applying linear temporal filtering to historical driving forces in the original coordinates and then transforming to the current step's weighted coordinates is equivalent to applying the same temporal filter in the current step's weighted coordinates. Memory is therefore accumulated and read out in the original coordinates as defined, and weighted by \(g\) and \(h\) only when \(R_t\) enters semi-orthogonalization. Estimation of \(\bar S\) and \(\bar C\) is detailed in Methods.

\subsection{Memory readout (③④⑤)}\label{ss:readout}

\paragraph{Three components of the driving force} We decompose the driving force on a hidden-layer weight matrix at step \(t\) as \(X_t=s_t+r_t+\xi_t\). Here \(s_t\) is the drift component, representing the sustained push of the mismatch-potential landscape on the medium and varying slowly across steps; \(r_t\) is the rebound component, an elastic restoring force caused by overshooting in the previous step's update whose direction flips sign at every step; and \(\xi_t\) is sampling noise arising from random fluctuations of the current mini-batch. Temporal memory must address all three simultaneously: tracking drift promptly, cancelling noise through multi-step averaging, and filtering out sign-flipping rebound. Because rebound depends only on the current weights, all micro-batch gradients within the same step contain the exact same rebound component, which cancels when taking differences between micro-batches; consequently, change ② cannot detect rebound from inter-micro-batch gradient differences, and rebound must be suppressed separately by the memory readout mechanisms (③--⑤) described here.

\paragraph{Rebound leakage through first-order relaxation} For a driving force \(X_t=(-1)^ta\) of constant amplitude that flips sign at every step, evolving under main-text equation~(3) gives, as \(t\to\infty\),
\begin{equation}
|M_t|\to\frac{1-\beta}{1+\beta}\,|a|;
\label{eq:leak}
\end{equation}
whereas for a constant driving force \(X_t=a\), \(M_t\to a\). In the language-model task, the steady-state rebound leakage ratio of the fast mode (\(\beta_f=0.85\)) is \(\rho_f=(1-\beta_f)/(1+\beta_f)\approx8.1\%\), whereas that of the slow mode (\(\beta_s=0.985\)) is only \(0.76\%\).

\paragraph{③ Gated readout} How much of the fast mode should be read out depends on the norm ratio between its residual rebound and its drift component. To estimate this ratio, PhysMuon computes two dimensionless quantities for each weight matrix at every step: \(o_t=\big(-\cos(X_t,M^f_{t-1})\big)_+\), measuring the anti-alignment between the current driving force and the previous step's fast mode, and \(d_t=\|M^s_{t-1}\|/\|X_t\|\), measuring the norm ratio of the previous step's slow mode to the current driving force. Consider a simplified model in which drift, rebound, and sampling noise are mutually orthogonal, both fast and slow modes are in steady state, and residual sampling noise in both modes as well as the tiny residual rebound in the slow mode are neglected. Under this model, \(\rho_fo_t/d_t\) depends solely on the norm ratio \(x\) of residual rebound to drift in the fast mode: when \(x\le\sqrt{\rho_f}\), its value is \(0\); when \(x>\sqrt{\rho_f}\) (that is, \(o_t>0\)), \(\rho_fo_t/d_t=\sqrt{1+x^2}-(1+\rho_f)/\sqrt{1+x^2}\), which is strictly increasing in \(x\). Accordingly, the gate sets the fast-mode readout weight to \(w_t=w_{\max}(1-\rho_fo_t/d_t)_+\) (Methods equation~(6)): \(w_t=w_{\max}\) in the absence of rebound, and \(w_t\) decreases as rebound strengthens; when \(\rho_fo_t/d_t\ge1\), \(w_t=0\), so the fast mode is not read out at all; for \(\beta_f=0.85\) in the language model this corresponds to \(x\approx1.32\), meaning that residual rebound in the fast mode exceeds the drift component by roughly \(30\%\). This readout rule has no explicit dependence on training step or learning rate. During the first few hundred steps after onset in language-model training, the gate reads out almost exclusively the slow mode; as the learning rate decays and rebound subsides, the fast-mode readout weight \(w_t\) gradually increases.

\paragraph{④ Tracking from step 0} If the fast and slow modes were initialized from the single-mode momentum only at the onset step \(T_{\mathrm{on}}\), the initial value of the slow mode would deviate from the state it should have accumulated on its own relaxation timescale. In the language-model task, for instance, the baseline momentum (\(\mu=0.95\)) averages gradients over roughly 19 steps, whereas the slow mode (\(\beta_s=0.985\)) averages over roughly 66 steps. In general, an initial-value deviation \(\delta M_0\) in the slow mode leaves an exact residual \(\beta_s^k\delta M_0\) after \(k\) steps of evolution: 125 steps after onset, \(0.985^{125}\approx15\%\) still remains, raising the language model's validation loss at that step by approximately \(0.02\) compared with tracking from step 0. Change ④ therefore lets both the fast and slow modes track the driving force continuously from step 0, with the onset step \(T_{\mathrm{on}}\) determining only when readout switches on; by the time readout is enabled at step 750, the influence of the step-0 initialization has decayed to \(0.985^{750}<10^{-4}\).

\paragraph{⑤ Fresh-share cap} In standard Muon implementations, what enters semi-orthogonalization is a mix \(R_t=(1-\mu)X_t+\mu M_t\) of instantaneous force and memory (a Nesterov-style mix with \(\mu=0.95\) on language models). The larger the slow-mode readout weight and the longer the effective memory duration, the more thoroughly sampling noise and rebound cancel in time averaging, and the smaller the memory norm \(\|M_t\|\). If the mixing coefficient \(\mu\) is held fixed, the share of the instantaneous branch in the sum of the two branch norms, \((1-\mu)\|X_t\|/\big((1-\mu)\|X_t\|+\mu\|M_t\|\big)\), hereafter termed the fresh share, rises accordingly, allowing more rebound to bypass memory and enter semi-orthogonalization directly. In the language-model task, when \(\mu=0.95\) is held fixed, the fresh share averages about \(33\%\) under a fixed readout ratio and rises to roughly \(60\%\) when only the slow mode is read out. Change ⑤ therefore sets the per-step mixing coefficient \(\mu_t\) according to Methods equation~(7) so that the fresh share never exceeds the cap \(\varphi\). Equation~(7) raises \(\mu_t\) to meet the cap only when the fresh share at the baseline value \(\mu\) would exceed \(\varphi\), keeping \(\mu_t=\mu\) otherwise; because the fresh share decreases monotonically with the mixing coefficient, this is the minimal adjustment satisfying the cap constraint.

\paragraph{Summary} Each of the five changes generalizes a minimal setting of standard Muon. Restricting ① to a single relaxation mode, setting \(g=h=I\) in ②, fixing the readout weight in ③, initializing from the baseline momentum at onset in ④, and setting \(\varphi=1\) in ⑤ respectively recover the baseline form of each change; disabling all five recovers standard Muon.

\section{Supplementary results on the language-model benchmark}\label{sn:lm}

\subsection{Two benchmark configurations}

The tuned Muon baseline is the official benchmark configuration (entry 36 in the results table), trained for 3,250 steps with a Muon learning rate of \(0.025\), weight decay of \(0.05\), and non-matrix parameters updated by AdamW~\cite{loshchilov2017adamw}. The record configuration is the leaderboard record at the time of our submission (PR \#328), which stacks SOAP~\cite{vyas2024soap}, sliding-window averaging of late-training weights (hereafter tail weight averaging), and several other components on top of Muon, with component definitions given in PR \#328. When using change ① to replace only the memory kernel in either configuration, all other code remains untouched and momentum is switched at step 1,000 to fast and slow relaxation modes (\(\beta_f=0.85\), \(\beta_s=0.98\), \(w=0.4385\)): both modes are initialized directly from the baseline momentum at step 1,000 and read out immediately, without enabling ④.

\subsection{Replacing only the memory kernel (①): summary and reruns}

On both the tuned Muon baseline and the record configuration, using change ① to replace only the momentum with fast and slow relaxation modes advances the first-crossing step, and tuning the fast-mode readout weight to \(w=1/6\) on tuned Muon further advances it to step 3,190; independently repeating the record configuration on an H100, or removing tail weight averaging to evaluate raw weights directly, preserves this advantage (Supplementary Table~\ref{stab:lmevidence}). In addition, rerunning the unmodified record configuration across 8 seeds on an A800 yields a first-crossing step of 2,695, within 5 steps of the 2,690 in the submission logs.

\begin{table}[tp]
\centering
\caption{Summary of results when replacing only the memory kernel (①). Lower first-crossing steps are better; `margin' is defined as \((3.28-\bar L)\sqrt n\) at the first-crossing step (\(\bar L\) is the mean validation loss across \(n\) seeds at that step), which must be at least \(0.004\) to meet the attainment criterion. Run logs are available in the public benchmark repository and the listed PRs.}
\label{stab:lmevidence}
\footnotesize
\setlength{\tabcolsep}{3.5pt}
\begin{tabular}{@{}lllc p{5.6cm}@{}}
\toprule
Configuration & Memory & First-crossing step & Seeds & Notes \\
\midrule
Tuned Muon & Momentum (Muon) & 3250 & 10 & Official baseline (entry 36), H100 logs \\
Tuned Muon & With ① & \textbf{3210} & 8 & A800, 40 steps earlier (PR \#340) \\
Tuned Muon & With ① (\(w=1/6\)) & \textbf{3190} & 8 & A800, fast-mode readout weight \(1/6\), 60 steps earlier \\
\midrule
Record & Momentum (Muon) & 2690 & 8 & Record at submission (PR \#328), A40 logs \\
Record & Momentum (A800 rerun) & 2695 & 8 & Same-GPU rerun of PR \#328 \\
Record & With ① (A800) & \textbf{2635} & 8 & 55 steps earlier (PR \#339), margin \(0.00419\) \\
Record & With ① (H100) & 2645 & 8 & Rerun on another GPU, margin \(0.00466\) \\
Record & With ① (combined) & 2635 & 16 & Margin \(0.00487\) \\
Record & Momentum, raw weights & 2735 & 8 & Without tail weight averaging \\
Record & With ①, raw weights & \textbf{2690} & 8 & Without tail weight averaging, still 45 steps earlier \\
Record & Momentum deepened to \(\bar n=30\) & 2775 & 8 & Same mean memory length as ①, A800 \\
\bottomrule
\end{tabular}
\end{table}

\subsection{The shape of the memory kernel is what matters}

In an exponential moving average with decay coefficient \(\beta\), the weighted mean lag of historical gradients from the current step is \(n(\beta)=\beta/(1-\beta)\); weighting the fast and slow modes by the readout ratio \(w\) gives the mean memory length of the composite kernel, \(\bar n=w\,n(\beta_f)+(1-w)\,n(\beta_s)\). On the record configuration, if single-mode momentum is deepened from step 1,000 onward to match the mean memory length of ① (\(\bar n=30\), i.e., \(\beta=30/31\), with all other settings unchanged across 8 seeds on an A800), the first-crossing step regresses to 2,775, 140 steps later than 2,635 with ① and also later than 2,690 of the unmodified record configuration; at step 2,635, by the benchmark's two-group comparison statistic, the difference in mean validation loss between this deepened single-mode group and ① divided by \(\sqrt{1/8+1/8}\) reaches \(0.0181\), well above the significance threshold \(0.004\). Fixing \(\bar n=19\) and testing two distinct parameter sets \((\beta_f,\beta_s,w)\) likewise outperforms single-mode momentum of the same \(\bar n\). Thus, what drives the improvement is the shape of the memory kernel rather than its mean memory length alone.

\subsection{Stepwise addition of changes}

Adding the five changes one by one to the tuned Muon baseline advances the first-crossing step at every stage (Supplementary Table~\ref{stab:ladder}); main-text Fig.~3a shows the first and last stages, Muon and PhysMuon. All stages were run on a single RTX 5090 using the same training script with only the corresponding switches toggled, 8 seeds per stage, recording validation loss every 5 steps after step 3,000. The stages with ① and with ①② retain the constants used in the ①-only experiments above, and together with Muon and ②-only form the \(2\times2\) factorial experiment below; starting from the stage adding ③, the constants switch to the PhysMuon values in Supplementary Table~\ref{stab:hparams}: the slow-mode decay coefficient of ① changes from \(0.98\) to \(0.985\), the readout weight switches from fixed \(0.4385\) to gated readout ③, and the onset step advances from step 1,000 to step 750.

\begin{table}[tp]
\centering
\caption{First-crossing step and mean validation loss at step 3,250 when adding the five changes sequentially to the tuned Muon baseline (8 seeds per stage, single RTX 5090). The lower three `Without ②' rows correspond to control configurations with ② disabled in the ③--⑤ sequence. \(\dagger\)~indicates switching the constants of ① to the PhysMuon values from that stage onward (see text).}
\label{stab:ladder}
\footnotesize
\begin{tabular}{@{}lcc@{}}
\toprule
Configuration & First-crossing step & Mean validation loss at step 3,250 \\
\midrule
Muon & Not reached & 3.27940 \\
Add ① & 3205 & 3.27683 \\
Add ①② & 3160 & 3.27316 \\
Add ①②③\(^\dagger\) & 3100 & 3.26986 \\
Add ①②③④ & 3095 & 3.26931 \\
Add ①②③④⑤ (PhysMuon) & 3085 & 3.26873 \\
\midrule
Without ②: ①③\(^\dagger\) & 3140 & 3.27284 \\
Without ②: ①③④ & 3130 & 3.27205 \\
Without ②: ①③④⑤ & 3120 & 3.27124 \\
\bottomrule
\end{tabular}
\end{table}

In the \(2\times2\) factorial experiment with 8 seeds per group, adding ② alone achieves a first-crossing step of 3,190 (PR \#370) and lowers the mean validation loss at step 3,250 by \(0.0040\) relative to the Muon baseline, decreasing in all 8 seeds (hereafter denoted 8/8); adding ② on top of ① lowers the validation loss by \(0.0037\) (8/8). The difference between the two reductions, namely the interaction term, is \(+0.0003\) with a 95\% confidence interval of \([-0.0002,+0.0007]\), showing that the gains from ① and ② are approximately additive. Furthermore, adding ⑤ on top of ①--④ lowers the mean validation loss at step 3,250 by an additional \(0.0006\) (8/8), and by \(0.0008\) (8/8) on the sequence without ②.

\subsection{302M model}

The 302M model scales GPT-2 to 24 layers and width 1,024, trained for 3,250 steps on a single RTX 5090 with 262,144 tokens per step and learning-rate multiplier \(0.7\) across 3 seeds (Supplementary Table~\ref{stab:hparams}). Full PhysMuon lowers the mean final validation loss from \(3.2395\) with Muon to \(3.2216\) (\(-0.0179\), 3/3 seeds). In a separate ablation script on the 302M model, adding ② alone lowers the mean validation loss at step 3,250 by \(0.013\) (3/3) relative to the Muon baseline, roughly three times its \(0.0040\) reduction on the 124M model.

\section{PhysMuon settings across tasks}\label{sn:tasks}

Supplementary Table~\ref{stab:hparams} lists the constants used across tasks. All five changes are enabled on language models; on interatomic potentials:
\begin{itemize}
\item All six interatomic-potential tasks use two-mode gated readout with changes ①--④, enabled at 30.77\% of total training steps;
\item ② uses the same constants as the language model, splitting each step's batch into 4 micro-batches on the DPA-3 tasks and using gradient differences across GPUs on DPA4-Mini (2 GPUs) and DPA4-Pro (4--16 GPUs);
\item ③ is enabled across all interatomic-potential tasks with cap \(w_{\max}=0.3\);
\item ④ is enabled on every task; ⑤ is used only for language models.
\end{itemize}

Because gradient rebound is weaker on interatomic potentials than in language models, the fresh-share cap ⑤ is omitted across all interatomic-potential tasks: omitting ⑤ lowers validation force MAE by \(2.0\%\) across all 6 seeds on regular-batch SPICE2 and avoids conflict with gated readout ③ on DPA4-Mini. Five of the six interatomic-potential tasks (DPA-3 on MPtrj, large-batch SPICE2, and ANI-1x, plus DPA4-Mini and DPA4-Pro) share identical two-mode constants \(\beta_f=0.90\), \(\beta_s=0.995\), and \(w_{\max}=0.3\), whereas the 8,000-step regular-batch SPICE2 task uses \(\beta_f=0.85\), \(\beta_s=0.98\), and \(w_{\max}=0.3\).

\begin{table}[tp]
\centering
\caption{Baseline and PhysMuon constants across tasks. `Onset step' is the step at which readout switches on, prior to which single-mode momentum with `Pre-onset' momentum is used. \(\beta_f\) and \(\beta_s\) are the decay coefficients of the fast and slow modes; \(w_{\max}\) is the fast-mode readout cap of gated readout ③. For ②, the exponent \(\lambda=1/2\), clipping interval \([1/2,2]\), and moving-average decay \(0.95\) are identical across all tasks; `Micro-batches' is the number of micro-batches (or GPUs for DPA4-Mini and DPA4-Pro) used by ②. `Baseline' gives the learning rate (or multiplier) and momentum of the reference Muon-family optimizer; `⑤' gives the fresh-share cap \(\varphi\), with `Off' indicating that ⑤ is disabled.}
\label{stab:hparams}
\footnotesize
\setlength{\tabcolsep}{2.2pt}
\begin{tabular}{@{}lllcccccc@{}}
\toprule
Task & Baseline & Onset step & Pre-onset & \(\beta_f\) & \(\beta_s\) & \(w_{\max}\) & Micro-batches & ⑤ \\
\midrule
GPT-2 124M & 0.025, 0.95 & 750 & 0.95 & 0.85 & 0.985 & ③, 0.6 & 8 & 0.25 \\
GPT-2 302M & 0.0175, 0.95 & 750 & 0.95 & 0.85 & 0.985 & ③, 0.6 & 16 & 0.25 \\
MPtrj (DPA-3) & \(\times8\), 0.98 & 6,154 & 0.95 & 0.90 & 0.995 & ③, 0.3 & 4 & Off \\
SPICE2 (DPA-3) & \(\times16\), 0.95 & 2,461 & 0.95 & 0.85 & 0.98 & ③, 0.3 & 4 & Off \\
SPICE2 large batch (DPA-3) & \(\times8\), 0.98 & 6,154 & 0.98 & 0.90 & 0.995 & ③, 0.3 & 4 & Off \\
ANI-1x (DPA-3) & \(\times8\), 0.98 & 6,154 & 0.98 & 0.90 & 0.995 & ③, 0.3 & 4 & Off \\
MPtrj (DPA4-Mini) & Official, 0.95 & 21,459 & 0.95 & 0.90 & 0.995 & ③, 0.3 & 2 & Off \\
MPtrj (DPA4-Pro) & Official, 0.95 & 615,400 & 0.95 & 0.90 & 0.995 & ③, 0.3 & 4--16 & Off \\
\bottomrule
\end{tabular}
\end{table}

\paragraph{Dataset versions and validation splits} Supplementary Table~\ref{stab:mlipdata} lists the dataset versions, split units, configuration counts, and fixed split identifiers for the three interatomic-potential datasets. Each dataset is partitioned at the material or molecule level so that all configurations of a held-out material or molecule belong only to the validation set. Reported validation errors across all interatomic-potential tasks therefore measure generalization to unseen materials on MPtrj and unseen molecules on SPICE2 and ANI-1x.

\begin{table}[tp]
\centering
\caption{Dataset versions and validation splits for the interatomic-potential tasks. All splits keep every configuration of a material or molecule in the same partition and evaluate generalization to unseen materials on MPtrj and unseen molecules on SPICE2 and ANI-1x. Split scripts and \texttt{mptrj\_validation\_ids.txt} are archived in the public repository.}
\label{stab:mlipdata}
\footnotesize
\setlength{\tabcolsep}{4pt}
\begin{tabular}{@{} >{\raggedright\arraybackslash}p{2.6cm} >{\raggedright\arraybackslash}p{4.0cm} >{\raggedright\arraybackslash}p{4.0cm} >{\raggedright\arraybackslash}p{4.5cm} @{}}
\toprule
Dataset (tasks) & Dataset version & Split unit and counts (train / val) & Fixed split identifier \\
\midrule
MPtrj\newline(DPA-3, DPA4-Mini, DPA4-Pro) & \texttt{2024-09-03-mp-trj}\newline(Matbench Discovery;\newline 145,923 materials) & By material (\texttt{mp-id});\newline 1,564,344 / 16,051\newline(1\% = 1,459 materials) & \texttt{random.Random(0)} over sorted \texttt{mp-id}s (\texttt{mptrj\_validation\_ids.txt}) \\[4pt]
SPICE2\newline(DPA-3 regular and large batch) & \texttt{InstaDeepAI/}\newline\texttt{SPICE2-curated} & By molecule;\newline 1,734,158 / 87,764 & Official source files (\texttt{SPICE2\_curated\_train.xyz} and \texttt{SPICE2\_curated\_val.xyz}) \\[4pt]
ANI-1x\newline(DPA-3) & \texttt{roitberg-group/ani1x}\newline(\texttt{ANI-1x-wB97X-631Gd.h5}) & By molecule (element sequence);\newline 4,852,676 / 103,329\newline(2\% = 56 of 3,114 molecules) & Deterministic SHA-1 hash of int64 element sequence (\texttt{\% 100 < 2}) \\
\bottomrule
\end{tabular}
\end{table}

\section{Supplementary results on interatomic potentials}\label{sn:mlip}

\subsection{Validation errors across tasks and per-seed results}

Supplementary Table~\ref{stab:mlipfull} lists validation force MAE, per-atom energy MAE, force RMSE across the five interatomic-potential tasks, and per-atom virial MAE on the two MPtrj tasks; Supplementary Table~\ref{stab:mlipseeds} reports per-seed validation force MAEs. All errors were evaluated using \texttt{dp test} in deepmd-kit, using exponential moving average weights for DPA4-Mini, and relative changes reported in the tables are means of seed-paired relative changes. Across every seed of all five tasks, PhysMuon achieves lower force, energy, and virial errors than the baseline, with force MAE reductions of \(7.4\%\)--\(20.9\%\) and energy MAE reductions of \(9.4\%\)--\(22.9\%\).

\subsection{Per-element force errors}

Supplementary Table~\ref{stab:elem} lists validation force MAE by chemical element on the 20,000-step SPICE2 large-batch task. Across all atoms, force MAE decreases from \(30.93\,\text{meV/\AA}\) with Muon to \(28.14\,\text{meV/\AA}\) with PhysMuon, a \(9.0\%\) reduction across all 6 seeds. Force MAE decreases consistently across all 12 elements by 8.1\%--16.8\%; for the 6 major elements accounting for more than 1\% of training atoms each (H, C, N, O, F, and S, totalling 98.6\%), reductions cluster tightly between 8.2\% and 9.8\%. PhysMuon therefore improves force predictions uniformly across both abundant and rare elements rather than on a few elements alone.

\subsection{DPA4-Pro settings}

Our DPA4-Pro run matches the official DPA-4.0.1-Pro-MPtrj architecture and training recipe, detailed side by side in Supplementary Table~\ref{stab:dpa4pro_recipe}. Both runs use the 22.8M-parameter model, a 2,000,000-step warmup-stable-decay cosine schedule from \(5\times10^{-4}\) to \(10^{-6}\), and MAE loss weights of 20, 20, and 5. Both also apply global gradient-norm clipping at 5.0 and a weight exponential moving average with decay 0.999. Whenever fewer than 16 GPUs were used, each \(k\)-GPU segment applied synchronized gradient accumulation with factor \(16/k\), including the 8-GPU segment with factor 2 from step 1.3M onward. Every optimizer step thus processed the official 16-GPU batch of 16 \texttt{filter:300} rank batches, so fewer steps correspond to proportionally less training data at matched effective batch size. Within this recipe, PhysMuon was enabled at 30.77\% of the 2-million total steps, step 615,400, with single-mode momentum \(0.95\) beforehand. The fast and slow modes of ① match DPA4-Mini at \(0.90\) and \(0.995\), with ③ capped at \(w_{\max}=0.3\), ④ enabled, and ⑤ disabled. Change ② starts at onset and estimates fluctuations from gradient differences across the 4--16 active GPUs.

During DPA4-Pro training, occasional near-contact crystal configurations produce heavy-tailed pre-clipping gradient norms exceeding the running median of roughly 12.5 by more than four orders of magnitude. Because deepmd-kit clips every step's global gradient norm to 5.0 before the optimizer step, these finite spikes enter the optimizer as finite clipped gradients with anomalous directions. To keep outliers out of the slow mode's 200-step memory, the slow mode skips any step whose pre-clipping norm exceeds 10 times the 1,000-step median or overflows float32. On such a step, the fast mode and weights update as usual on the finite clipped gradient. Change ② computes squared inter-rank gradient differences before clipping, so extreme finite spikes can also overflow float32 in those squared differences. From step 1.42M onward, ②'s fluctuation estimate therefore skips any step flagged by the spike gate or producing non-finite pre-clipping squared differences and keeps its previous value.

The main result is taken from step 1.465M. Evaluating the official public DPA-4.0.1-Pro-MPtrj checkpoint with the same relaxation and phono3py scripts reproduces its leaderboard scores. Under the Matbench Discovery workflow, PhysMuon checkpoints achieve F1, \(\kappa_{\mathrm{SRME}}\), \(\mathrm{RMSD}_{\mathrm{AUC}}\), and CPS of \(0.8607\), \(0.2106\), \(0.7719\), and \(0.8654\) at step 900k; \(0.8650\), \(0.2003\), \(0.7735\), and \(0.8698\) at step 1.1M; \(0.8686\), \(0.1948\), \(0.7739\), and \(0.8727\) at step 1.3M; and \(0.8712\), \(0.1860\), \(0.7734\), and \(0.8757\) at step 1.465M, compared with \(0.8615\), \(0.2114\), \(0.7651\), and \(0.8650\) for the official 2.0M-step leaderboard checkpoint. Supplementary Table~\ref{stab:matbench} compares our 1.465M-step checkpoint with the top five MPtrj-compliant models on the Matbench Discovery leaderboard.

\begin{table}[tp]
\centering
\caption{Validation errors across the five interatomic-potential tasks. Each entry gives mean error for baseline \(\to\) PhysMuon, with mean seed-paired relative change and number of improved seeds in parentheses. Forces are in meV/\AA; energies and virials are in meV per atom. Virials are available only on the two MPtrj tasks, as molecular datasets do not carry virial labels.}
\label{stab:mlipfull}
\footnotesize
\setlength{\tabcolsep}{3pt}
\begin{tabular}{@{}l l *{4}{>{\centering\arraybackslash}p{2.35cm}}@{}}
\toprule
Task & Baseline & Force MAE & Energy MAE & Force RMSE & Virial MAE \\
\midrule
MPtrj (DPA-3) & Muon & 123.5 $\to$ 97.6\newline($-20.9\%$, 4/4) & 41.8 $\to$ 32.2\newline($-22.9\%$, 4/4) & 295.5 $\to$ 215.0\newline($-26.9\%$, 4/4) & 58.4 $\to$ 49.5\newline($-15.2\%$, 4/4) \\
SPICE2 (DPA-3) & Muon & 69.6 $\to$ 63.7\newline($-8.3\%$, 6/6) & 18.5 $\to$ 16.2\newline($-12.0\%$, 6/6) & 120.9 $\to$ 111.9\newline($-7.4\%$, 6/6) & -- \\
SPICE2 large batch (DPA-3) & Muon & 30.9 $\to$ 28.1\newline($-9.0\%$, 6/6) & 5.9 $\to$ 4.9\newline($-16.6\%$, 6/6) & 57.6 $\to$ 52.8\newline($-8.3\%$, 6/6) & -- \\
ANI-1x (DPA-3) & Muon & 46.9 $\to$ 42.2\newline($-10.2\%$, 4/4) & 7.5 $\to$ 5.8\newline($-22.8\%$, 4/4) & 85.5 $\to$ 76.6\newline($-10.3\%$, 4/4) & -- \\
MPtrj (DPA4-Mini) & HybridMuon & 46.7 $\to$ 43.3\newline($-7.4\%$, 3/3) & 18.5 $\to$ 16.8\newline($-9.4\%$, 3/3) & 142.9 $\to$ 121.0\newline($-14.4\%$, 3/3) & 33.7 $\to$ 32.3\newline($-4.2\%$, 3/3) \\
\bottomrule
\end{tabular}
\end{table}

\begin{table}[tp]
\centering
\caption{Per-seed validation force MAE (meV/\AA) across the five interatomic-potential tasks. Within each task, the baseline and PhysMuon in the same column use identical random seeds and training steps.}
\label{stab:mlipseeds}
\footnotesize
\setlength{\tabcolsep}{4pt}
\begin{tabular}{@{}llcccccc@{}}
\toprule
Task & Optimizer & seed 0 & seed 1 & seed 2 & seed 3 & seed 4 & seed 5 \\
\midrule
MPtrj (DPA-3) & Muon & 128.63 & 126.35 & 121.20 & 117.71 &  &  \\
 & PhysMuon & 97.52 & 95.11 & 100.33 & 97.33 &  &  \\[2pt]
SPICE2 (DPA-3) & Muon & 70.19 & 68.91 & 71.42 & 68.54 & 65.24 & 73.16 \\
 & PhysMuon & 63.27 & 60.84 & 66.39 & 63.68 & 63.57 & 64.58 \\[2pt]
SPICE2 large batch (DPA-3) & Muon & 30.58 & 30.31 & 31.54 & 30.67 & 30.50 & 31.97 \\
 & PhysMuon & 27.50 & 28.08 & 28.55 & 27.68 & 28.02 & 28.98 \\[2pt]
ANI-1x (DPA-3) & Muon & 46.88 & 46.19 & 47.26 & 47.37 &  &  \\
 & PhysMuon & 42.16 & 41.38 & 43.05 & 42.01 &  &  \\[2pt]
MPtrj (DPA4-Mini) & HybridMuon & 47.38 & 46.23 & 46.56 &  &  &  \\
 & PhysMuon & 43.54 & 43.48 & 42.81 &  &  &  \\
\bottomrule
\end{tabular}
\end{table}

\begin{table}[tp]
\centering
\caption{Per-element validation force MAE (meV/\AA, mean across 6 seeds) on the 20,000-step SPICE2 large-batch task. `Training share' is the percentage of total training-set atoms belonging to that element; the last column reports the mean seed-paired relative change of PhysMuon versus Muon, computed using the same convention as Supplementary Table~\ref{stab:mlipfull}.}
\label{stab:elem}
\footnotesize
\setlength{\tabcolsep}{6pt}
\begin{tabular}{@{}lrrrr@{}}
\toprule
Element & Training share & Muon & PhysMuon & Change (PhysMuon vs.\ Muon) \\
\midrule
H & 45.6\% & 16.0 & 14.5 & $-8.9\%$ \\
B & 0.13\% & 63.1 & 57.3 & $-9.2\%$ \\
C & 37.0\% & 41.1 & 37.5 & $-8.9\%$ \\
N & 6.57\% & 58.2 & 52.9 & $-9.1\%$ \\
O & 7.20\% & 40.2 & 36.3 & $-9.8\%$ \\
F & 1.03\% & 27.4 & 25.1 & $-8.2\%$ \\
Si & 0.18\% & 77.3 & 70.0 & $-9.4\%$ \\
P & 0.09\% & 145.3 & 124.5 & $-14.2\%$ \\
S & 1.22\% & 64.9 & 59.0 & $-9.1\%$ \\
Cl & 0.72\% & 24.8 & 22.8 & $-8.1\%$ \\
Br & 0.21\% & 28.0 & 24.7 & $-11.6\%$ \\
I & 0.05\% & 40.4 & 33.3 & $-16.8\%$ \\
\midrule
All atoms & 100\% & 30.93 & 28.14 & $-9.0\%$ \\
\bottomrule
\end{tabular}
\end{table}

\begin{table}[tp]
\centering
\caption{Side-by-side training recipe comparison between the official Matbench Discovery DPA-4.0.1-Pro-MPtrj entry and our PhysMuon run.}
\label{stab:dpa4pro_recipe}
\footnotesize
\setlength{\tabcolsep}{4.5pt}
\begin{tabular}{@{} >{\raggedright\arraybackslash}p{3.0cm} >{\raggedright\arraybackslash}p{6.0cm} >{\raggedright\arraybackslash}p{6.2cm} @{}}
\toprule
Setting & Official DPA-4.0.1-Pro-MPtrj & Our PhysMuon run \\
\midrule
Architecture & DPA-4.0.1-Pro (22.8M parameters; bfloat16 mixed-precision descriptor, TF32) & Identical (verbatim \texttt{model\_params}) \\[4pt]
Data and split & MPtrj & MPtrj with 1\% \texttt{mp-id} validation holdout (1,564,344 train / 16,051 val frames) \\[4pt]
Effective global batch & 16 GPUs \(\times\) \texttt{filter:300} per GPU & 16 effective ranks \(\times\) \texttt{filter:300} (4--16 GPUs with accumulation \(16/k\); 8 GPUs \(\times\) 2 from 1.3M steps) \\[4pt]
LR schedule & 2.0M-step WSD cosine (\(5\times10^{-4}\to10^{-6}\), 5,000 warmup, 0.65 decay ratio) & Identical \\[4pt]
Loss and clipping & MAE (20, 20, 5 energy, force, virial weights), global gradient-norm clip \(5.0\) & Identical \\[4pt]
Weight EMA & Exponential moving average (decay \(0.999\)); evaluated on EMA weights & Identical \\[4pt]
Optimizer & HybridMuon (\texttt{slice}, Gram, Magma-lite, wd \(10^{-3}\), \(\mu=0.95\)) & HybridMuon with PhysMuon enabled at step 615,400 (① 0.90/0.995, ②, ③ \(w_{\max}=0.3\), ④; ⑤ off); \(\mu=0.95\) beforehand \\[4pt]
Spike and overflow handling & Global gradient-norm clipping (\(5.0\)) & Global gradient-norm clipping (\(5.0\)); slow mode skips pre-clipping norm \(>10\times\) 1,000-step median or float32 norm overflow; ② skips gated and overflow steps from 1.42M \\[4pt]
Evaluated checkpoint & 2.0M-step leaderboard checkpoint & 1.465M-step EMA checkpoint (checkpoints from 900k onward all \(\ge\) official CPS) \\
\bottomrule
\end{tabular}
\end{table}

\begin{table}[tp]
\centering
\caption{Comparison with the top five MPtrj-compliant universal interatomic potentials on the Matbench Discovery leaderboard (retrieved November 2026; ranked by CPS among the 26 MPtrj-compliant entries reporting all four metrics). Reference values are taken from official Matbench Discovery model records, with unique-prototype F1 computed to four decimals from reported confusion counts and CPS from \(0.5\,\mathrm{F1}+0.4\,(1-\kappa_{\mathrm{SRME}}/2)+0.1\,\mathrm{RMSD}_{\mathrm{AUC}}\). Higher is better for CPS, F1, and \(\mathrm{RMSD}_{\mathrm{AUC}}\); lower is better for \(\kappa_{\mathrm{SRME}}\). Bold marks the best value in each column.}
\label{stab:matbench}
\footnotesize
\setlength{\tabcolsep}{5pt}
\begin{tabular}{@{}lcccc@{}}
\toprule
Model (MPtrj-compliant) & CPS & F1 & \(\kappa_{\mathrm{SRME}}\) & \(\mathrm{RMSD}_{\mathrm{AUC}}\) \\
\midrule
DPA-4.0.1-Pro-MPtrj (PhysMuon, 1.465M) & \textbf{0.8757} & \textbf{0.8712} & \textbf{0.1860} & \textbf{0.7734} \\
DPA-4.0.1-Pro-MPtrj (HybridMuon, 2.0M) & 0.8650 & 0.8615 & 0.2114 & 0.7651 \\
EquiformerV3+DeNS-MP & 0.8549 & 0.8680 & 0.2753 & 0.7590 \\
DPA-4.0-Pro-MPtrj & 0.8529 & 0.8503 & 0.2414 & 0.7604 \\
eSEN-30M-MP & 0.8233 & 0.8362 & 0.3398 & 0.7322 \\
MatRIS-10M-MP & 0.8027 & 0.8516 & 0.4892 & 0.7480 \\
\bottomrule
\end{tabular}
\end{table}

\section{Supplementary results on molecular dynamics}\label{sn:md}

\subsection{Molecules and DFT calculations}

Molecules for MD simulations were selected from the respective validation sets: we first chose a set of target atom counts, 10, 12, 13, 15, 16, 17, 18, 21, 22, 23, 25, and 27 for ANI-1x and 18, 24, 30, 36, 42, 48, 54, and 60 for SPICE2, and for each atom count selected the single-molecule configuration without stretched bonds (all bond lengths within 1.15 times the sum of covalent radii) that had the smallest root-mean-square atomic force as the starting structure. The 8 molecules selected from SPICE2 are listed in Supplementary Table~\ref{stab:mdmol}.

Reference DFT values were computed with PySCF~\cite{sun2020pyscf}: MD snapshots used \(\omega\)B97x/6-31G(d) (Cartesian d functions, density fitting), whose forces differ from dataset labels by an average of \(1.6\,\text{meV/\AA}\) across 12 validation configurations; bond-stretching scans used SPICE2's native \(\omega\)B97M-D3(BJ)/def2-TZVPPD level of theory, differing from dataset labels by an average of \(0.4\,\text{meV/\AA}\) across the 3 starting structures, and were evaluated with spin-unrestricted open-shell calculations taking the lowest-energy solution as the bond dissociates.

\subsection{Regular tests and energy conservation}

In regular constant-temperature tests, PhysMuon models incurred zero instabilities across all 576 trajectories, whereas Muon models incurred one instability in a \(1500\,\mathrm K\) ANI-1x trajectory of C\(_{9}\)H\(_{12}\)O\(_{4}\) at \(28.28\,\mathrm{ps}\).

In the energy-conservation tests run for \(40\,\mathrm{ps}\) at \(300\,\mathrm K\) with the thermostat turned off (96 trajectories per optimizer, 48 each on ANI-1x and SPICE2), the mean absolute total-energy drift on ANI-1x is \(0.0005\,\mathrm{meV}\) per atom per picosecond for both optimizers, whereas on SPICE2 it decreases from \(0.0005\) to \(0.0004\,\mathrm{meV}\) per atom per picosecond with PhysMuon. Counting individual trajectories across both datasets, the number of trajectories with drift exceeding \(10^{-3}\,\mathrm{meV}\) per atom per picosecond drops from 12 for Muon models to 9 for PhysMuon models, with the maximum drift decreasing from \(0.0028\) to \(0.0016\,\mathrm{meV}\) per atom per picosecond.

\subsection{High-temperature stress tests}

Supplementary Table~\ref{stab:mdstress} summarizes the high-temperature stress tests at \(1200\)--\(2100\,\mathrm K\) for the 20,000-step SPICE2 models trained with Muon and PhysMuon, and Supplementary Table~\ref{stab:mdstats} reports seed-level statistical comparisons. Because all 8 test molecules share the same trained weights within each seed, statistical inference is performed over the \(n=6\) independently trained model seeds. Across seeds 0--5 (56 trajectories per seed and temperature), unstable trajectory counts for Muon versus PhysMuon are \([0,0,0,1,1,0]\) versus \([0,0,1,0,0,0]\) at \(1200\,\mathrm K\), \([2,2,2,5,3,1]\) versus \([1,1,1,0,2,1]\) at \(1500\,\mathrm K\), \([9,8,13,30,24,4]\) versus \([7,10,14,4,8,15]\) at \(1800\,\mathrm K\), and \([23,32,44,48,48,29]\) versus \([20,31,33,20,22,38]\) at \(2100\,\mathrm K\). PhysMuon lowers instability counts in 5 of 6 seeds with 1 tie at \(1500\,\mathrm K\) and in 5 of 6 seeds at \(2100\,\mathrm K\). The instability rate drops by factors of \(2.5\) at \(1500\,\mathrm K\) (exact seed-cluster 95\% confidence interval \(1.46\)--\(6.85\)) and \(1.7\) at \(2100\,\mathrm K\) (95\% confidence interval \(1.01\)--\(3.01\)), and the 42-ps restricted mean survival time (RMST) at \(2100\,\mathrm K\) increases by \(6.0\,\mathrm{ps}\) (95\% confidence interval \(0.5\)--\(11.4\,\mathrm{ps}\)). Across all four temperatures combined, total unstable trajectories drop from 329 to 229. Per-molecule counts appear in Supplementary Table~\ref{stab:mdmol}, and PhysMuon incurs fewer instabilities on all 8 molecules at \(2100\,\mathrm K\).

Linearly interpolating the SPICE2 instability fraction as a function of temperature shows that PhysMuon models reach Muon's instability levels at \(1200\), \(1500\), and \(1800\,\mathrm K\) only at \(1260\,\mathrm K\) (exact seed-cluster bootstrap 95\% confidence interval \(1200\)--\(1500\,\mathrm K\)), \(1552\,\mathrm K\) (\(1514\)--\(1638\,\mathrm K\)), and \(1885\,\mathrm K\) (\(1685\)--\(2060\,\mathrm K\)), an increase of \(50\)--\(85\,\mathrm K\) in tolerated temperature.

\subsection{Bond-stretching scans and bond-dissociation temperatures}

We selected three covalent bonds frequently broken in SPICE2 stress tests (Si--Cl, C--H, and Si--C, in C\(_5\)H\(_{10}\)Cl\(_2\)Si, C\(_8\)H\(_{13}\)NO\(_2\), and C\(_{17}\)H\(_{18}\)Si, respectively), pulled the dissociating atom or group step-by-step along the bond axis from the starting configuration while holding all other atoms fixed, and compared model-predicted energy rises against DFT references. Within \(\pm0.2\,\text{\AA}\) of the equilibrium bond length, both model families deviate from DFT by less than \(0.05\,\mathrm{eV}\). At extensions of \(1\)--\(3\,\text{\AA}\), where interatomic potentials soften relative to DFT, PhysMuon models track DFT more closely across all three bonds: the ratio of model-predicted energy rise to DFT increases from \(65.9\%\), \(60.0\%\), and \(83.9\%\) for Muon to \(69.8\%\), \(63.4\%\), and \(89.7\%\) for PhysMuon on Si--Cl, C--H, and Si--C, respectively (Supplementary Table~\ref{stab:bondscan}), a \(5.7\%\)--\(6.9\%\) lift in stretching energy. Averaged over all 9 non-zero scan points (\(-0.2\), \(0.2\), \(0.5\), \(0.7\), \(1.0\), \(1.5\), \(2.0\), \(2.5\), and \(3.0\,\text{\AA}\)), PhysMuon models lower full-curve energy MAE by \(8.6\%\)--\(16.4\%\): \(0.784\) versus \(0.878\,\mathrm{eV}\) for Si--Cl, \(0.751\) versus \(0.821\,\mathrm{eV}\) for C--H, and \(0.380\) versus \(0.454\,\mathrm{eV}\) for Si--C.

The more a bond-stretching energy is underestimated, the more easily thermal motion breaks the bond: in the Boltzmann factor \(\exp(-E/k_BT)\), scaling energy by \(\lambda\) shifts the dissociation temperature by the same factor. Fitting all SPICE2 stress-test instability events across \(1200\)--\(2100\,\mathrm K\) to a shared-prefactor Arrhenius relation yields an effective energy barrier for PhysMuon models that is \(1.065\) times that of Muon models (molecule-stratified bootstrap 95\% confidence interval \(1.025\)--\(1.105\); for all C--H cleavages, which account for 211 versus 115 events, \(\lambda=1.089\)), matching the \(5.7\%\)--\(6.9\%\) lift measured independently in the rigid DFT bond-stretching scans (Supplementary Table~\ref{stab:bondscan}).

\begin{table}[tp]
\centering
\caption{High-temperature stress-test results for 20,000-step SPICE2 models. At each temperature, 336 trajectories (8 molecules \(\times\) 6 seeds, 7 trajectories per group) were run per optimizer for up to \(42\,\mathrm{ps}\). `Remaining stable' columns give the percentage of trajectories that have not undergone dissociative instability up to that time; `42-ps RMST' is the 42-ps restricted mean survival time, with trajectories remaining stable throughout the run capped at \(42\,\mathrm{ps}\).}
\label{stab:mdstress}
\footnotesize
\setlength{\tabcolsep}{5pt}
\begin{tabular}{@{}llrrrrrr@{}}
\toprule
 & & & \multicolumn{4}{c}{Remaining stable (\%)} & \\
\cmidrule(lr){4-7}
Temperature & Optimizer & Unstable trajectories & \(10\,\mathrm{ps}\) & \(20\,\mathrm{ps}\) & \(30\,\mathrm{ps}\) & \(42\,\mathrm{ps}\) & 42-ps RMST (ps) \\
\midrule
\(1200\,\mathrm K\) & Muon & 2 (0.6\%) & 100 & 100 & 100 & 99 & 41.9 \\
 & PhysMuon & 1 (0.3\%) & 100 & 100 & 100 & 100 & 41.9 \\[2pt]
\(1500\,\mathrm K\) & Muon & 15 (4.5\%) & 98 & 98 & 97 & 96 & 41.0 \\
 & PhysMuon & 6 (1.8\%) & 99 & 99 & 99 & 98 & 41.6 \\[2pt]
\(1800\,\mathrm K\) & Muon & 88 (26.2\%) & 88 & 82 & 78 & 74 & 35.0 \\
 & PhysMuon & 58 (17.3\%) & 93 & 90 & 87 & 83 & 37.8 \\[2pt]
\(2100\,\mathrm K\) & Muon & 224 (66.7\%) & 71 & 51 & 41 & 33 & 23.4 \\
 & PhysMuon & 164 (48.8\%) & 80 & 69 & 59 & 51 & 29.4 \\
\bottomrule
\end{tabular}
\end{table}

\begin{table}[tp]
\centering
\caption{Seed-level statistical comparison between PhysMuon and Muon in SPICE2 high-temperature stress tests. Because all 8 molecules share the same trained weights within each seed, paired sign-flip permutation tests are enumerated exactly over all \(2^6=64\) sign assignments of the \(n=6\) independently trained model seeds (two-sided; minimum attainable \(p=2/64\approx0.031\) with 6 non-tied seeds and \(p=4/64=0.0625\) when 1 seed ties). `Seeds (fewer/more)' reports how many of the 6 seeds incur fewer or more unstable trajectories under PhysMuon than under Muon (remaining seeds tie); `Pairs (fewer/more)' lists the descriptive count across the 48 molecule-seed pairs. `Instability-rate ratio' is the ratio of unstable trajectories per unit simulation time (Muon / PhysMuon). Values in parentheses are exact 95\% confidence intervals from exhaustive seed-cluster bootstrap resampling over all \(6^6\) seed draws jointly across all 8 molecules; the upper bound is \(\infty\) at \(1200\,\mathrm K\) because some resampled seed sets contain zero PhysMuon instabilities.}
\label{stab:mdstats}
\footnotesize
\setlength{\tabcolsep}{5pt}
\begin{tabular}{@{}lcccc@{}}
\toprule
Temperature & Seeds (fewer/more) & Seed-level \(p\) & Pairs (fewer/more) & Instability-rate ratio \\
\midrule
\(1200\,\mathrm K\) & 2/1 & 1.00 & 2/1 & 2.0 (0.0--\(\infty\)) \\
\(1500\,\mathrm K\) & 5/0 & 0.0625 & 9/4 & 2.5 (1.5--6.9) \\
\(1800\,\mathrm K\) & 3/3 & 0.44 & 19/9 & 1.6 (0.6--4.1) \\
\(2100\,\mathrm K\) & 5/1 & 0.16 & 26/9 & 1.7 (1.01--3.0) \\
\midrule
All four combined & 4/1 & 0.31 & 30/15 & 1.5 (0.9--2.7) \\
\bottomrule
\end{tabular}
\end{table}

\begin{table}[tp]
\centering
\caption{Unstable trajectory counts by molecule in SPICE2 high-temperature stress tests (Muon / PhysMuon). At each temperature, 42 trajectories (6 seeds \(\times\) 7 trajectories per seed) were run per molecule and optimizer.}
\label{stab:mdmol}
\footnotesize
\setlength{\tabcolsep}{6pt}
\begin{tabular}{@{}llcccc@{}}
\toprule
Molecule & Formula & \(1200\,\mathrm K\) & \(1500\,\mathrm K\) & \(1800\,\mathrm K\) & \(2100\,\mathrm K\) \\
\midrule
m0 & C\(_{5}\)H\(_{10}\)Cl\(_{2}\)Si & 0 / 0 & 0 / 0 & 6 / 1 & 13 / 12 \\
m1 & C\(_{8}\)H\(_{13}\)NO\(_{2}\) & 0 / 0 & 6 / 1 & 20 / 19 & 38 / 31 \\
m2 & C\(_{10}\)H\(_{17}\)NOS & 0 / 0 & 0 / 0 & 2 / 0 & 18 / 13 \\
m3 & C\(_{17}\)H\(_{18}\)Si & 1 / 1 & 7 / 4 & 26 / 27 & 42 / 41 \\
m4 & C\(_{14}\)H\(_{23}\)NO\(_{3}\)Si & 0 / 0 & 0 / 0 & 6 / 1 & 25 / 7 \\
m5 & C\(_{14}\)H\(_{32}\)N\(_{2}\) & 0 / 0 & 0 / 0 & 3 / 1 & 33 / 17 \\
m6 & C\(_{21}\)H\(_{25}\)N\(_{3}\)O\(_{5}\) & 0 / 0 & 1 / 1 & 10 / 4 & 25 / 19 \\
m7 & C\(_{25}\)H\(_{27}\)N\(_{5}\)O\(_{3}\) & 1 / 0 & 1 / 0 & 15 / 5 & 30 / 24 \\
\bottomrule
\end{tabular}
\end{table}

\begin{table}[tp]
\centering
\caption{Rigid DFT bond-stretching scans on 20,000-step SPICE2 models. The `Muon' and `PhysMuon' columns report the model-predicted energy rise as a percentage of the DFT reference across extensions of \(1\)--\(3\,\text{\AA}\) (five points: \(1\), \(1.5\), \(2\), \(2.5\), and \(3\,\text{\AA}\)), averaged across 6 seeds. `Energy gain' is the percentage by which PhysMuon's predicted energy rise exceeds Muon's over \(1\)--\(3\,\text{\AA}\); `Full-curve MAE' lists Muon \(\to\) PhysMuon energy MAE averaged over all 9 non-zero scan points (\(-0.2\) to \(3.0\,\text{\AA}\)), with relative change in parentheses.}
\label{stab:bondscan}
\footnotesize
\setlength{\tabcolsep}{5pt}
\begin{tabular}{@{}llcccc@{}}
\toprule
 & & \multicolumn{2}{c}{Fraction of DFT energy rise} & & \\
\cmidrule(lr){3-4}
Bond & Molecule & Muon & PhysMuon & Energy gain & Full-curve MAE (eV) \\
\midrule
Si--Cl & m0 & 65.9\% & 69.8\% & \(+6.0\%\) & \(0.878 \to 0.784\) (\(-10.7\%\)) \\
C--H & m1 & 60.0\% & 63.4\% & \(+5.7\%\) & \(0.821 \to 0.751\) (\(-8.6\%\)) \\
Si--C & m3 & 83.9\% & 89.7\% & \(+6.9\%\) & \(0.454 \to 0.380\) (\(-16.4\%\)) \\
\bottomrule
\end{tabular}
\end{table}

\section{Read-only probe: optimal memory duration grows during training}\label{sn:probe}

\subsection{Measured quantity}

Momentum tracks the gradient while filtering out sampling noise. At each step, the gradient decomposes into two parts: a drift component that persists across steps and varies slowly during training, and a batch-specific sampling-noise component (batch noise). Momentum averages recent gradients over a sliding window: a longer window suppresses batch noise more thoroughly but responds more slowly to changes in drift. The optimal window length is governed by the ratio of drift strength to noise strength~\cite{kalman1960}: stronger noise favours a longer window, whereas faster drift variation favours a shorter window. Based on this ratio, the probe outputs at each step the reading \(K^\star=\sqrt{\hat D/\hat T}\), where \(\hat D\) and \(\hat T\) are estimates of drift strength and batch-noise strength, respectively; a smaller \(K^\star\) corresponds to a longer optimal memory. Supplementary Fig.~\ref{sfig:probe} plots the proxy for optimal memory length corresponding to \(K^\star\),
\begin{equation}
n^\star=\frac{\sqrt{1+4/(K^\star)^2}-1}{2},
\end{equation}
which is the mean lag in steps of an exponential moving average whose optimal gain, namely the weight assigned to each new gradient, is set from the drift-to-noise ratio; when \(K^\star\ll1\), the optimal gain is approximately \(K^\star\) and \(n^\star\approx1/K^\star-1/2\). Under the physical picture of a responsive medium, this optimal memory duration should grow as training progresses.

\subsection{Measurement protocol}

The probe ran alongside unmodified single-mode momentum training of the record configuration from scratch across 8 independent trajectories (seeds 0--7); it only read and never altered model parameters, gradients, or optimizer states.

Batch noise was estimated via half-batch splitting: at each optimization step, the current data batch was split into two equal halves and gradients were computed on each half separately. Denoting the two half-batch gradients by \(G_A\) and \(G_B\), the full-batch gradient is \(G=(G_A+G_B)/2\), whereas in the half-batch difference \(N=(G_A-G_B)/2\) the shared drift component cancels, leaving only batch noise.

Measurements were taken along the principal directions actually updated by momentum. At each step, we performed a singular value decomposition of the matrix \(\widetilde M\) entering semi-orthogonalization, which is the whitened momentum in the record configuration,
\begin{equation*}
\widetilde M=\sum_{i=1}^{r}\sigma_iu_iv_i^{\mathsf T}\qquad(\sigma_1\ge\sigma_2\ge\cdots\ge\sigma_r),
\end{equation*}
and projected \(G\) and \(N\) onto each singular direction via \(g_i=u_i^{\mathsf T}Gv_i\) and \(\nu_i=u_i^{\mathsf T}Nv_i\). All directions were partitioned by relative rank \(i/r\) at boundaries \(0\), \(0.05\), \(0.15\), \(0.35\), \(0.65\), and \(1\) into five groups, where Group 1 corresponds to the strongest 5\% of directions. Within group \(b\), we computed group means \(s_b=\frac{1}{n_b}\sum_{i\in b}g_i\) and \(T_b=\frac{1}{n_b}\sum_{i\in b}\nu_i^2\), where \(n_b\) is the number of directions in group \(b\).

Each group of each matrix maintained two exponential moving averages with decay coefficient \(0.99\): \(\hat T_b\) tracking \(T_b\), and \(E_b\) tracking the squared difference between consecutive steps, \(\bigl(s_b(t)-s_b(t-1)\bigr)^2\). Subtracting the batch-noise contributions of the two consecutive steps from \(E_b\) yields the drift strength \(\hat D_b\) and the corresponding \(K^\star_b\):
\begin{equation}
\hat D_b=\max\bigl(E_b-2\hat T_b/n_b,\ 10^{-12}\bigr),\qquad K^\star_b=\sqrt{\hat D_b/\hat T_b}.
\end{equation}
With 5 groups across each of the 72 hidden-layer matrices, this forms 360 measurement units, each producing one \(K^\star\) reading per step; the first 250 steps served as a warm-up period for the moving averages and were excluded from statistics. Because the five relative-rank groups have unequal widths \(n_b\) (spanning 5\%--35\% of singular directions), \(K^\star_b\) differs from the single-direction optimal gain by a group-dependent scale factor; quantitative growth is therefore interpreted strictly through within-group relative changes between an early window (steps 700--1,500) and a late window (steps 2,200--2,700), where constant group width cancels, while cross-group comparisons assess only rank consistency between windows.

\subsection{Results}

\begin{figure}[tp]
\centering
\includegraphics[width=\textwidth]{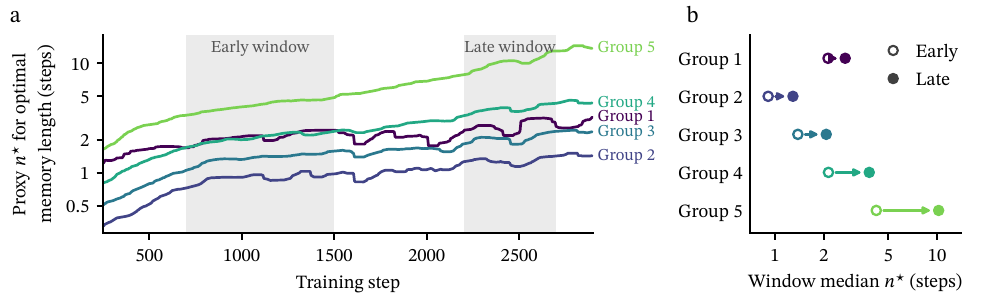}
\caption{Optimal memory duration measured by the read-only probe (unmodified training of the record configuration, seed 0). (a) Each curve takes the median of \(n^\star\) for a direction group across the 72 hidden-layer matrices at each step, followed by a 101-step rolling median along training steps. Groups 1 to 5 correspond to the strongest 5\%, 5\%--15\%, 15\%--35\%, 35\%--65\%, and weakest 35\% of singular directions, respectively. The vertical axis is logarithmic; the first 250 warm-up steps are omitted, and grey shaded bands mark the early and late time windows. (b) Unsmoothed window-median \(n^\star\) of each group in the two windows (open circles, early window; filled circles, late window).}
\label{sfig:probe}
\end{figure}

\begin{figure}[tp]
\centering
\includegraphics[width=0.69\textwidth]{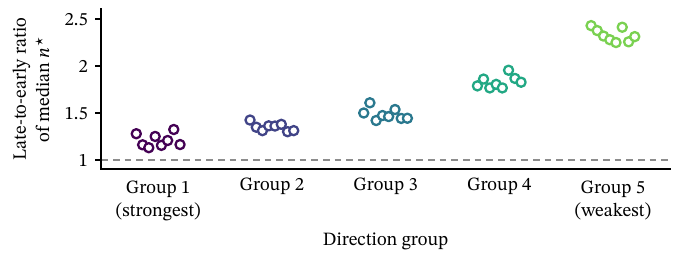}
\caption{Reproduction of the growth in optimal memory duration across 8 independent training trajectories. Each point gives the ratio of late-window to early-window median \(n^\star\) for one direction group in a single trajectory, with the 8 points within each group slightly jittered horizontally; the horizontal dashed line marks a ratio of 1.}
\label{sfig:probe8}
\end{figure}

Supplementary Fig.~\ref{sfig:probe} shows the measurements for seed 0, where each curve is the median of \(n^\star\) across the 72 hidden-layer matrices for one direction group. From the early window to the late window, \(n^\star\) rises consistently across all five groups: the strongest group increases from 2.1 to 2.7 steps, while the weakest group increases from 4.2 to 10.2 steps. The ordering of the five groups is preserved between the early and late windows (Supplementary Fig.~\ref{sfig:probe}b), and the ranking across all 360 measurement units is highly consistent with a Spearman rank correlation of 0.89; meanwhile, the interquartile range of \(\log K^\star\) across units widens from 0.743 to 0.940, indicating increasing scale separation among units. This growth of \(n^\star\) during training is reproduced across all 8 independent trajectories (Supplementary Fig.~\ref{sfig:probe8}): all 40 late-to-early ratios of median \(n^\star\) exceed 1 (ranging from 1.13 to 2.43), with weaker singular directions exhibiting larger increases in optimal memory duration. This phenomenon is consistent with the dynamical slowing down reported in comparative studies of deep-network training and structural glass aging~\cite{kerrwinter2024glassy}.

\subsection{Relation to gated readout}

When the probe estimates drift from differences between consecutive steps, step-flipping rebound with a period of roughly 2 steps is also included in the difference. To separate rebound from drift, we performed two diagnostic training runs on the tuned Muon baseline, one with fixed-readout ① only and one with gated-readout ①③ across 2 seeds each, decomposing per-step gradient components along the top 16 principal directions of the slow mode into drift, rebound, and batch noise, and recomputing \(K^\star\) using two-step differences so that period-2 rebound cancels. In the early window, rebound power is 2.6--6.7 times the drift-increment power, dropping to 1.0--1.7 times in the late window; from the early to the late window, \(K^\star\) computed from one-step and two-step differences decreases to 0.42--0.53 and 0.73--0.94 of its early-window value, respectively. Thus, even after removing rebound, the drift-estimated optimal memory duration still grows by 6\%--37\% during training under the standard decaying learning-rate schedule, where shrinking per-step displacements and intrinsic landscape slowing both contribute; at the same time, strong early-training rebound requires additional memory depth to filter out rebound. Gated readout in change ③ satisfies both requirements: early in training it uses a fast-mode weight of only 0.05--0.1, reading out almost exclusively the slow mode to suppress rebound, and later in training, as rebound subsides, it returns to the optimal memory depth estimated from drift.

\section{Computational overhead}\label{sn:cost}

On the language-model benchmark (GPT-2 124M), GPU memory footprint and per-step wall-clock time were measured on a single RTX 5090 and verified on a single A800. Running the same script for 40 steps with Muon and PhysMuon, with the onset step of PhysMuon set to step 10 so that all code paths are exercised, and averaging per-step time over steps 20--40 after compilation, per-step time increases only from \(2.620\,\mathrm s\) to \(2.660\)--\(2.682\,\mathrm s\) (\(+1.5\%\)--\(2.4\%\)) on the RTX 5090, and by only \(2.9\%\) on the A800. Persistent memory increases from \(1.468\) to \(2.119\,\mathrm{GiB}\), with optimizer-state memory increasing from \(0.749\) to \(1.383\,\mathrm{GiB}\), while peak per-step memory increases from \(28.96\) to \(29.93\,\mathrm{GiB}\).

Across complete 3,250-step training runs on identical RTX 5090 hardware, the 8 seeds of PhysMuon took an average total time of \(8709\,\mathrm s\), \(2.1\%\) more than the \(8528\,\mathrm s\) mean of 4 standalone timing runs of the unmodified Muon baseline (the 8-seed factorial Muon baseline script averaged \(8660\,\mathrm s\), with 5 of 8 seeds reaching \(3.28\) in an average of \(8550\,\mathrm s\)). All eight PhysMuon runs reached a validation loss of \(3.28\) within 3,250 steps, as did three of the four standalone timing runs with unmodified Muon. Among these successful runs, the mean wall-clock time to first reach the target was \(8{,}228\,\mathrm{s}\) for PhysMuon and \(8{,}485\,\mathrm{s}\) for Muon, approximately \(3\%\) shorter with PhysMuon.

In interatomic-potential tasks, when training DPA4-Pro at the production batch size, PhysMuon uses only about \(175\,\mathrm{MiB}\) more memory per GPU than Muon; the additional memory comes from optimizer states and is independent of batch size.

\bibliographystyle{unsrtnat}
\bibliography{references}